\documentclass[letterpaper]{article} 
\usepackage{aaai2027} 
\nocopyright
\usepackage[hyphens]{url}  
\usepackage{graphicx} 
\usepackage{natbib}  
\usepackage{caption} 
\usepackage{algorithm}
\usepackage{algorithmic}
\usepackage{multirow}
\usepackage{amssymb}

\usepackage{newfloat}
\usepackage{listings}
\DeclareCaptionStyle{ruled}{labelfont=normalfont,labelsep=colon,strut=off} 
\floatstyle{ruled}
\newfloat{listing}{tb}{lst}{}
\floatname{listing}{Listing}

\usepackage{booktabs}

\title{Beyond Illumination: A Conditional Mutual Information-Guided Network for Low-Light Image Enhancement} 
\author{
    Ya-nan Guan\textsuperscript{\rm 1, \rm 2},
    Shaonan Zhang\textsuperscript{\rm 3},
    Tao Dai\textsuperscript{\rm 2, \rm 3}\corresponding,
    Tianqu Zhuang\textsuperscript{\rm 2},
    Yongchao Qiao\textsuperscript{\rm 4},
    Zhensen Chen\textsuperscript{\rm 5},
    Shu-Tao Xia\textsuperscript{\rm 2},
    Hang Guo\textsuperscript{\rm 2},
}
\affiliations{
    \textsuperscript{\rm 1}College of Artificial Intelligence, Nankai University, Tianjin, China\\
    \textsuperscript{\rm 2}Tsinghua Shenzhen International Graduate School, Tsinghua University, Shenzhen, China\\
    \textsuperscript{\rm 3}College of Computer Science and Software Engineering, Shenzhen University, China\\
    \textsuperscript{\rm 4}School of Systems Science and Engineering, Sun Yat-sen University, Guangzhou, China\\
    \textsuperscript{\rm 5}Jimei University, Xiamen, China\\
    guanyanan@mail.nankai.edu.cn, 13414561874@163.com, zhuangtq23@mails.tsinghua.edu.cn, \\
    qiaoyc2023@163.com, cshguo@gmail.com, 202561000132@jmu.edu.cn, \\
    daitao@szu.edu.cn, xiast@sz.tsinghua.edu.cn
}

\begin{document}

\maketitle

\begin{abstract}

Low-light image enhancement (LLIE) seeks to restore structural fidelity, natural color rendition, and proper exposure from images captured under inadequate lighting conditions. Recent state-of-the-art approaches, such as CIDNet, adopt a dual-branch architecture comprising a chrominance (HV) branch and an intensity (I) branch to separately model decoupled chromatic and luminance information within the HVI color space. However, these methods overlook the mutual interaction between intensity and chrominance components, which inherently limits their representational capacity and leads to suboptimal enhancement performance.
To address this limitation, we propose the Conditional Mutual Information-Guided Network (CMIG-Net), which leverages conditional mutual information as a principled metric to quantitatively assess the contribution of chrominance features conditioned on the available intensity information. In particular, we design a Conditional Mutual Information Calibration (CMIC) module that generates a conditional information map, enabling region-adaptive recalibration of chrominance representations according to local illumination statistics. Furthermore, we introduce a Dynamic Dual-branch Information Restoration (D2IR) module, which adaptively governs bidirectional information flow between the intensity and chrominance branches, guided by both the conditional prior and the instantaneous restoration state.
Extensive experiments on paired LLIE benchmarks demonstrate that CMIG-Net consistently outperforms CIDNet, achieving up to a 0.619 dB gain in PSNR, with a 0.382 dB improvement specifically on the challenging Sony-Total-Dark dataset.
\end{abstract}


\section{Introduction}\label{sec:Introduction}


Low-light image enhancement (LLIE) aims to recover high-quality image content from severely degraded observations captured under suboptimal illumination conditions. Early deep learning methods \cite{lore2017llnet, guo2020zero} primarily formulate this task as a direct end-to-end mapping in the RGB domain. Subsequent representative approaches draw on Retinex theory \cite{zhang2019kindling, Chen2018Retinex, cai2023retinexformer, wu2022uretinex}, shifting the focus toward illumination map estimation and adjustment. These advances have led to a broad consensus in LLIE that illumination restoration plays a dominant role, since it determines the fundamental visibility, global contrast, and structural boundaries of the restored image. 

\begin{figure}[!tb]
    \centering
    \includegraphics[width=0.48\textwidth]{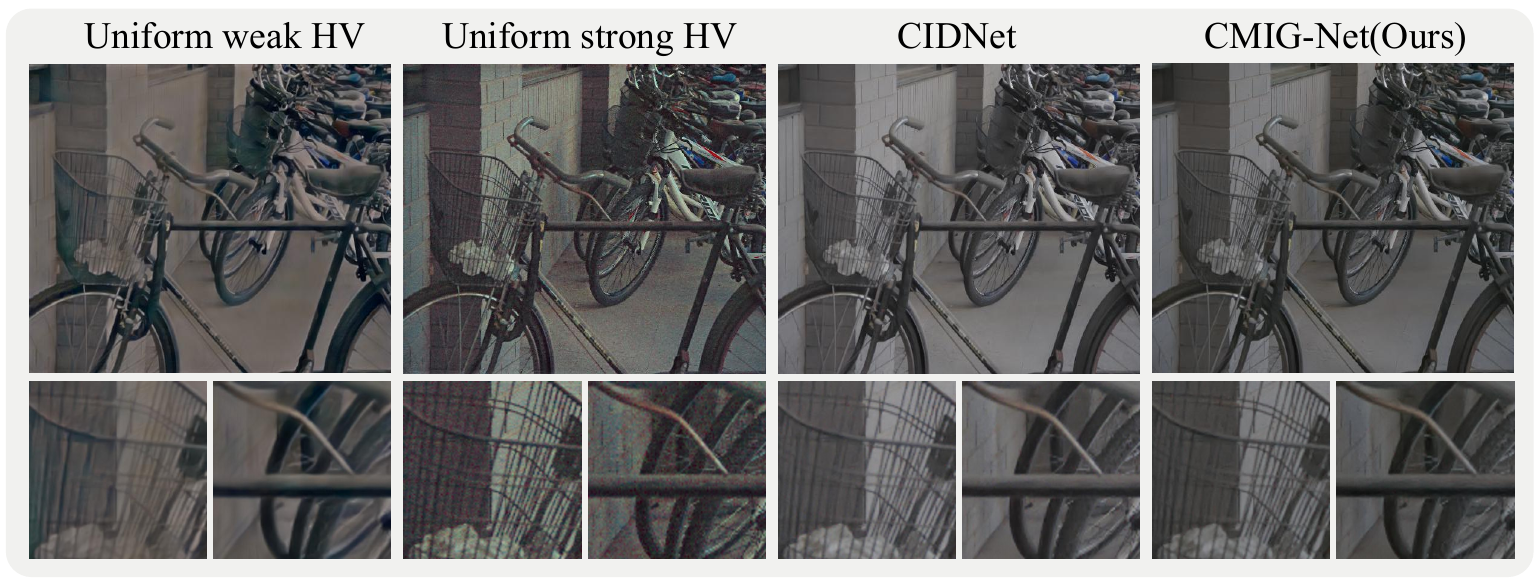}
    \caption{Effects of HV chrominance injection. 
Weak HV injection suppresses noise, while strong injection enhances details. CIDNet without HV injection fails to restore details, while our CMIG-Net with HV injection produces sharper results.}
    \label{fig:motivation}
\end{figure}

Following such an illumination-dominant perspective,
recent HVI-based methods \cite{xu2025hsvnet, yan2025hvi, xu2026iclr} have achieved remarkable performance by disentangling input RGB images into intensity (I space) and chrominance (HV space) components in the HVI space.
However, such HVI-based methods employ an HV branch and an intensity branch to model decoupled color and brightness information, respectively, but fail to exploit the interactions between the intensity and chrominance components \cite{yan2025hvi, xu2026iclr}, resulting in suboptimal performance, as Figure \ref{fig:motivation}).
In fact, chrominance cues are also helpful for detail recovery. As shown in Figure \ref{fig:motivation}, weak HV injection suppresses noise, while strong HV injection enhances structural details.

\begin{figure}[!tb]
    \centering
    \includegraphics[width=0.46\textwidth]{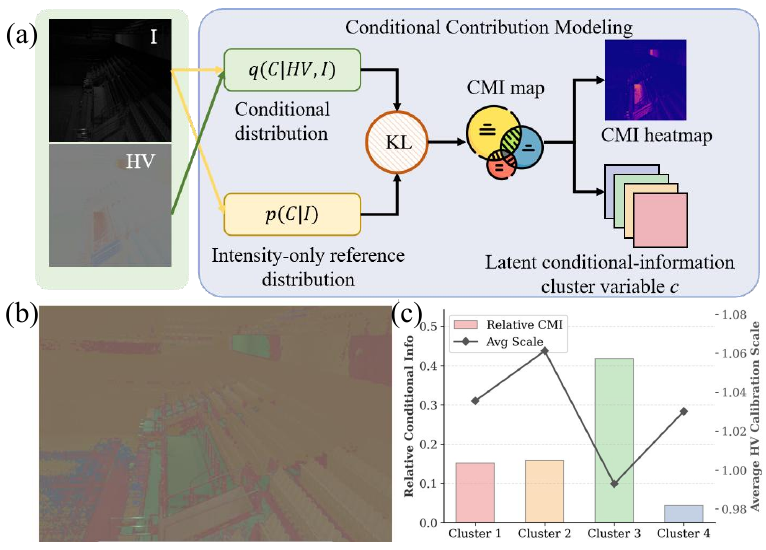}
    \vspace{-2mm}
    \caption{Empirical analysis of chrominance contribution in the HVI space. (a) Conditional contribution modeling. (b) Spatial clusters. (c) Cluster-wise contribution statistics.}
    \label{fig:observation}
\end{figure}

To further investigate the effects of chrominance information,
we conduct an empirical analysis of information interaction in the HVI space. As shown in Figure \ref{fig:observation}(a), an image is decomposed into intensity and chrominance components. We estimate a conditional distribution from both components and a reference distribution from intensity alone. Their discrepancy measures how much additional information chrominance provides given intensity. Figure \ref {fig:observation}(b) and (c) further show that this contribution varies across regions. Different clusters exhibit distinct relative contribution strengths and calibration scales. Therefore, a fixed injection strength cannot accommodate all regions.

Motivated by the above observations, we propose the Conditional Mutual Information-Guided Network (CMIG-Net) for low-light image enhancement, which explicitly quantifies and dynamically exploits the additional chrominance information. First, we propose a Conditional Mutual Information Calibration (CMIC) module to construct a spatial conditional-information map from the discrepancy between the complete HVI conditional distribution and the intensity-only reference distribution. This map quantifies the additional contribution of HV across different regions and is subsequently used to calibrate the input chrominance representation. Second, our CMIG-Net, composed of Dynamic Dual-branch Information Restoration (D2IR) modules, employs the conditional-information map as a conditional prior throughout the restoration stages. Within each D2IR module, bidirectional Conditional Information-Guided Attention (CIGA) jointly considers this prior and the current restoration states of the intensity and chrominance branches to dynamically regulate information injection along both I-to-HV and HV-to-I. Consequently, our CMIG-Net transforms input-level contribution quantification into spatially and stage-adaptive information interaction, thereby exploiting informative complementary cues while suppressing redundant or unreliable information. Overall, we make three key contributions:
\begin{itemize}
    \item We propose a conditional mutual information (CMI)-based metric to estimate which chrominance components provide useful cues beyond intensity and use it to guide region-adaptive chrominance calibration.
    
    \item We propose the Conditional Mutual Information-Guided Network (CMIG-Net), which consists of a Conditional Mutual Information Calibration (CMIC) module and Dynamic Dual-branch Information Restoration (D2IR) modules.
     CMIC quantifies and calibrates the spatial contribution of chrominance, while D2IR dynamically regulates dual-branch information injection according to the current restoration state.
    
    \item Extensive experiments demonstrate the effectiveness of our CMIG-Net in terms of quantitative and qualitative metrics. Compared with the state-of-the-art CIDNet, our method obtains PSNR gains of 0.6 dB, 0.3 dB, and 0.8 dB on the LOLv1, LOLv2-Real, and LOLv2-Synthetic datasets, respectively.
\end{itemize}

\section{Related Work}\label{sec:Related Work}

\subsection{Low-Light Image Enhancement (LLIE)}\label{sec:Low-Light Image Enhancement}

Low-light image enhancement aims to restore the visibility, structural details, and color fidelity of images captured under insufficient illumination. Early methods, including LLNet \cite{lore2017llnet}, EnlightenGAN \cite{jiang2021enlightengan}, and Zero-DCE \cite{guo2020zero}, directly learn enhancement mappings in the RGB space. LLFlow \cite{wang2022low}, LLFormer \cite{wang2023ultra}, and diffusion-based models improve restoration by exploiting probabilistic distributions, long-range dependencies, and generative priors, respectively. Another line of research uses Retinex theory to decompose illumination and reflectance. RetinexNet \cite{Chen2018Retinex}, KinD \cite{zhang2019kindling}, URetinex-Net \cite{wu2022uretinex}, and Retinexformer \cite{cai2023retinexformer} improve illumination estimation and content restoration through convolutional networks, deep unfolding, and Transformers. CIDNet \cite{yan2025hvi} and ICLR \cite{xu2026iclr} further disentangle intensity and chrominance in the HVI space and perform joint restoration through dual-branch interactions. Existing methods primarily rely on attention, modulation, or statistical constraints to fuse chrominance information implicitly. They do not explicitly quantify the additional contribution of chrominance given intensity.

\subsection{Mutual Information Modeling in Low-Level Vision}\label{sec:Mutual Information Modeling in Low-Level Vision}

Mutual information measures the statistical dependence between random variables and has been widely used in visual representation learning and low-level vision \cite{cover1999elements, hjelm2018learning}. MITNet \cite{shen2023mutual} minimizes mutual information to reduce cross-stage feature redundancy and learn complementary representations in the spatial and frequency domains. In low-light visual perception, mutual information has also served as a metric for evaluating the correlation between enhanced results and input content \cite{xu2023mutual}. Conditional mutual information (CMI) further characterizes the incremental dependence between two variables given a third variable \cite{wyner1978definition}. Yang et al. use conditional mutual information to characterize intra-class compactness in the output distributions of classification networks and propose a corresponding constrained learning framework \cite{yang2025conditional}. Guo et al. learn degradation-independent representations for camera ISPs by minimizing conditional incremental information \cite{guo2024learning}. Related studies also apply CMI to multimodal information fusion \cite{li2025infobridge} and diffusion-based inverse problem solving \cite{hamidi2025conditional}. LLIE needs to assess the incremental contribution of chrominance given intensity. CMI naturally captures this dependence and is therefore well suited to guiding region-adaptive chrominance exploitation.

\begin{figure*}[!tb]
    \centering
    \includegraphics[width=0.90\textwidth]
    {\detokenize{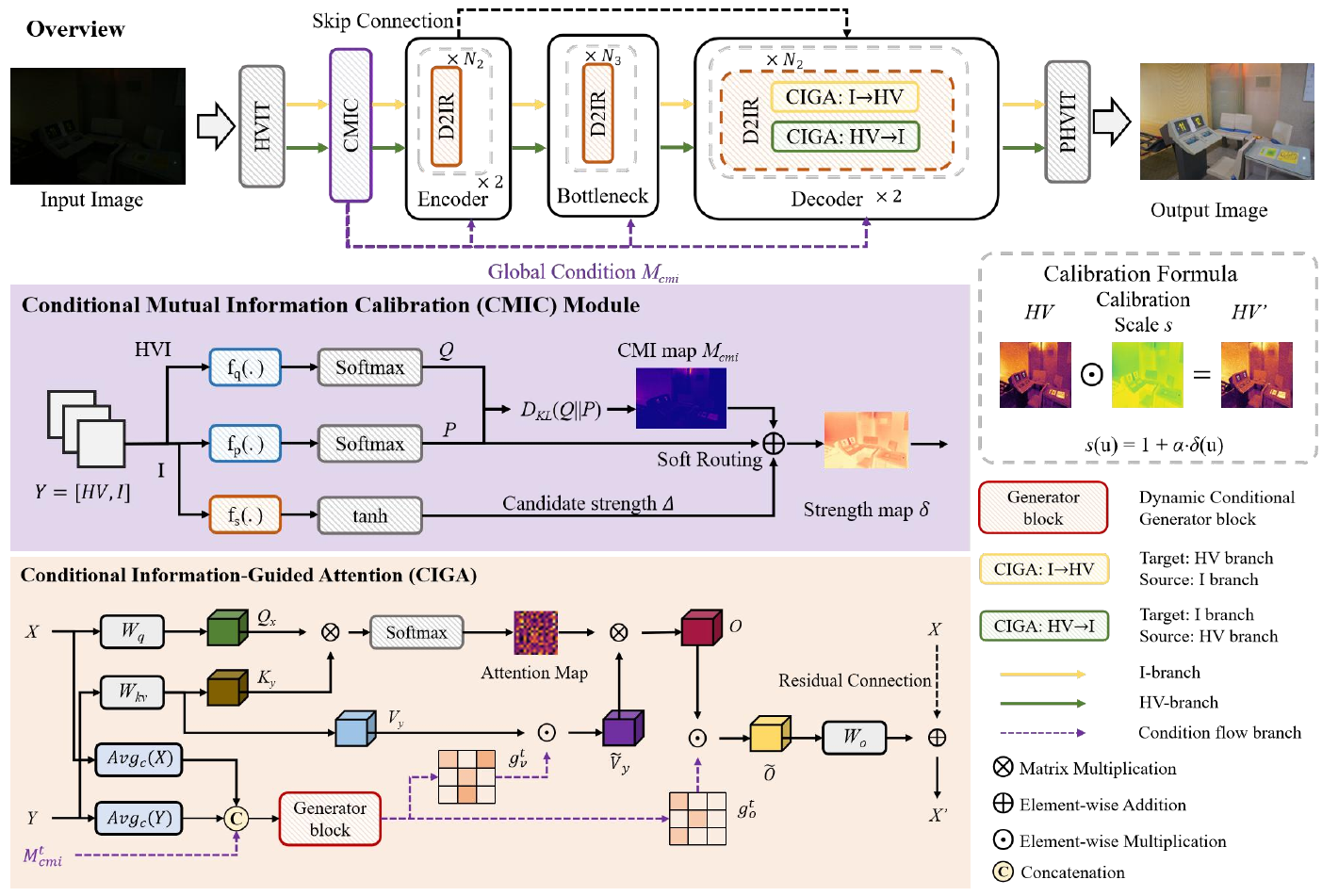}}
    \caption{Overall architecture of our CMIG-Net, which mainly consists of
    the Conditional Mutual Information Calibration (CMIC) module and the Dynamic Dual-branch Information Restoration (D2IR) module based on Conditional Information-Guided Attention (CIGA).
    CMIC generates the global conditional CMI map and calibrates chrominance. D2IR performs dual-branch restoration, where CIGA dynamically regulates cross-branch information injection.}
    \label{fig:overall_architecture}
\end{figure*}

\section{Preliminaries and Motivation}\label{sec:Preliminaries}

\subsection{Problem Formulation in the HVI Space}\label{sec:HVI Space}

In the RGB space, intensity degradation and color distortion are entangled, making their contributions hard to separate. We therefore adopt the HVI representation \cite{yan2025hvi}, which decomposes an image into intensity $I$ and chrominance $HV$. $I$ describes exposure and structure, while $HV$ provides complementary color and detail cues but may also contain noise and artifacts. To quantify the additional useful contribution of $HV$ given $I$, we introduce a latent conditional-information cluster variable $C$. If adding $HV$ changes the cluster distribution to a greater extent than that inferred from $I$ alone, then $HV$ contributes more strongly. We model this contribution as the conditional mutual information (CMI) $\mathcal{I}(HV; C \mid I)$.

\subsection{Derivation of Conditional Mutual Information}\label{sec:Learnable Proxies and Variational Derivation}

Since $HVI=[HV,I]$, the chain rule of mutual information gives

\begin{equation}
\mathcal{I}(HV; C \mid I)=\mathcal{I}(HVI; C)-\mathcal{I}(I; C).
\label{eq:01}
\end{equation}

A similar conditional mutual information decomposition has been used in information bottleneck analysis \cite{zhuang2025stealthy}.
Equivalently, the CMI in Eq. \ref{eq:01} can be written as
\begin{equation}
\mathcal{I}(HV; C \mid I)=\mathrm{E}_{HVI,C}\left[\log P(C \mid HVI)-\log P(C \mid I)\right].
\label{eq:02}
\end{equation}

Eq. \ref{eq:02} shows that the conditional contribution of $HV$ given $I$ can be characterized by the difference in log-likelihoods between two conditional predictions. This discrepancy captures the additional contribution of $HV$ to distinguishing the latent conditional-information cluster variable when $I$ is given. Since the true conditional distributions are unavailable, we introduce two learnable proxy distributions

\begin{equation}
\begin{array}{c}
q_{\phi}(C \mid HV,I) \approx P(C \mid HVI), p_{\psi}(C \mid I) \approx P(C \mid I).
\end{array}
\label{eq:03}
\end{equation}

$q_{\phi}(C|HV,I)$ is predicted from the complete HVI representation and thus incorporates information from both $I$ and $HV$. It is regarded as the full-information conditional distribution. In contrast, $p_{\psi}(C|I)$ is predicted solely from the $I$ component and serves as the intensity-only reference distribution. Their discrepancy reflects how the distribution of $C$ changes after $HV$ is introduced relative to the distribution inferred from $I$ alone.

Based on these proxy distributions, we construct a pixel-wise approximation of CMI. At each spatial location, both $q_{\phi}$ and $p_{\psi}$ output a distribution over the $K$ latent clusters, and the CMI map is
\begin{equation}
\begin{array}{c}
M_{\mathrm{cmi}}
=
\sum_{c=1}^{K}
q_{\phi}(c \mid HV,I)
\log
\frac{q_{\phi}(c \mid HV,I)}
     {p_{\psi}(c \mid I)}\\[2pt]
= D_{\mathrm{KL}}
\!\left(
q_{\phi}(C \mid HV,I)
\mathbin{\|}
p_{\psi}(C \mid I)
\right).
\label{eq:04}
\end{array}
\end{equation}

\section{Methodology}\label{sec:Methodology}

Motivated by the empirical observations in Figure \ref{fig:observation} and the derivation in Sec. \ref{sec:Preliminaries}, we design CMIG-Net. As shown in Figure \ref{fig:overall_architecture}, the input is mapped to the HVI space by HVIT and mapped back to RGB by PHVIT after restoration. CMIG-Net consists of the Conditional Mutual Information Calibration (CMIC) module and the Dynamic Dual-branch Information Restoration (D2IR) module, whose core interaction unit is Conditional Information-Guided Attention (CIGA).

\subsection{Conditional Mutual Information Calibration Module}\label{sec:CMIC}

Based on the above derivation, CMIC performs two functions. First, it estimates the additional contribution of $HV$ to the latent conditional-information cluster variable $C$ given $I$. Second, it adaptively calibrates $HV$ according to this conditional contribution. We therefore construct a lightweight CMIC module. It takes the HVI representation $Y=[HV,I]$ as input and produces the calibrated representation $Y'=[HV',I]$, as illustrated in Figure \ref{fig:overall_architecture}.
For the HVI representation, the $q$-branch predicts the full-information conditional distribution, whereas the $p$-branch predicts the intensity-only reference distribution solely from the $I$ component:

\begin{equation}
\begin{array}{c}
Q=q_{\phi}(C \mid HV,I)
=\mathrm{Softmax}\!\left(f_q([HV,I])\right), \\[2pt]
P=p_{\psi}(C \mid I)
=\mathrm{Softmax}\!\left(f_p(I)\right).
\end{array}
\label{eq:06}
\end{equation}

In Eq. \ref{eq:06}, $f_q(\cdot)$ and $f_p(\cdot)$ have the same lightweight architecture. At each spatial location, $Q$ denotes the probability distribution over the $K$ states of $C$ after both $I$ and $HV$ are observed. In contrast, $P$ denotes the reference distribution of $C$ inferred solely from the $I$ component.
We compute the spatial CMI map from the discrepancy between $Q$ and $P$. At spatial location $u$, CMI is approximated as

\begin{equation}
M_{\mathrm{cmi}}(u)
=
\sum_{c=1}^{K}
Q_c(u)
\log
\frac{Q_c(u)}
     {P_c(u)+\epsilon}.
\label{eq:07}
\end{equation}
In Eq. \ref{eq:07}, $K$ is the number of states of $C$. A larger $M_{\mathrm{cmi}}(u)$ indicates a stronger additional contribution of $HV$ to $C$ at that location, whereas a smaller value indicates a weaker or less reliable contribution.
Based on the conditional distribution, CMIC further performs cluster-conditioned calibration of $HV$. First, $I$ is used to generate an $HV$ calibration candidate for each state of $C$: $\Delta=\tanh\!\left(f_s(I)\right)$, $\Delta\in\mathbb{R}^{K\times H \times W}$, where $f_s$ outputs the calibration response for each state of $C$. The candidates are then softly routed by $Q$, yielding the effective calibration strength:

\begin{equation}
\delta(u)
=
\sum_{c=1}^{K}
Q_c(u)\Delta_c(u).
\label{eq:08}
\end{equation}

Using the calibration strength in Eq. \ref{eq:08}, the calibration scale is defined as $s(u)=1+\alpha\cdot\delta(u)$, where $\alpha$ is the maximum modulation amplitude. CMIC obtains the calibrated representation through location-wise scale modulation, $HV'(u)=HV(u)\odot s(u)$, while leaving the $I$ component unchanged. Finally, it outputs $Y'=[HV',I]$.

\begin{figure*}[t]
\centering
\scriptsize
\setlength{\tabcolsep}{4.0pt}
\renewcommand{\arraystretch}{0.8}
\resizebox{\textwidth}{!}{
\begin{tabular}{l|cc|ccc|ccc|ccc}
\hline
\textbf{Methods}
& \multicolumn{2}{c|}{\textbf{Complexity}}
& \multicolumn{3}{c|}{\textbf{LOLv1}}
& \multicolumn{3}{c|}{\textbf{LOLv2-Real}}
& \multicolumn{3}{c}{\textbf{LOLv2-Synthetic}} \\
& Params/M & FLOPs/G
& PSNR$\uparrow$ & SSIM$\uparrow$ & LPIPS$\downarrow$
& PSNR$\uparrow$ & SSIM$\uparrow$ & LPIPS$\downarrow$
& PSNR$\uparrow$ & SSIM$\uparrow$ & LPIPS$\downarrow$ \\
\hline

RetinexNet \cite{Chen2018Retinex} &
0.840 & 584.470 &
18.915 & 0.427 & 0.470 &
16.097 & 0.401 & 0.543 &
17.137 & 0.762 & 0.255 \\

KinD \cite{zhang2019kindling} &
8.020 & 34.990 &
23.018 & 0.843 & 0.156 &
17.544 & 0.669 & 0.375 &
18.320 & 0.796 & 0.252 \\

Zero-DCE \cite{guo2020zero} &
0.075 & 4.830 &
21.880 & 0.640 & 0.335 &
16.059 & 0.580 & 0.313 &
17.712 & 0.815 & 0.169 \\

LLFlow \cite{wang2022low} &
17.420 & 358.400 &
24.998 & 0.871 & 0.117 &
17.433 & 0.831 & 0.176 &
24.807 & 0.919 & 0.067 \\

SNR-Aware \cite{xu2022snr} &
4.010 & 26.350 &
26.716 & 0.851 & 0.152 &
21.480 & 0.849 & 0.163 &
24.140 & 0.928 & 0.056 \\

PairLIE \cite{fu2023learning} &
0.330 & 20.810 &
23.526 & 0.755 & 0.248 &
19.885 & 0.778 & 0.317 &
19.074 & 0.794 & 0.230 \\

LLFormer \cite{wang2023ultra} &
24.550 & 22.520 &
25.758 & 0.823 & 0.167 &
20.056 & 0.792 & 0.211 &
24.038 & 0.909 & 0.066 \\

Retinexformer \cite{cai2023retinexformer} &
1.530 & 15.850 &
27.140 & 0.850 & 0.129 &
22.794 & 0.840 & 0.171 &
25.670 & 0.930 & 0.059 \\

GSAD \cite{hou2023global} &
17.360 & 442.020 &
27.605 & 0.876 & 0.092 &
20.153 & 0.846 & \textcolor{red}{0.113} &
24.472 & 0.929 & 0.051 \\

CWNet \cite{zhang2025cwnet}&
1.230 & 11.300 &
26.381 & 0.866 & 0.119 &
21.647 & 0.860 & 0.136 &
25.501 & 0.936 & \textcolor{blue}{0.044} \\

URWKV \cite{xu2025urwkv}&
2.250 & 18.340 &
26.091 & 0.869 & 0.104 &
23.180 & 0.863 & 0.132 &
\textcolor{blue}{26.284} &
\textcolor{red}{0.944} &
0.046 \\

CIDNet \cite{yan2025hvi}&
1.880 & 7.570 &
\textcolor{blue}{27.890} &
\textcolor{red}{0.879} &
\textcolor{blue}{0.080} &
23.905 & 0.866 &
\textcolor{blue}{0.122} &
25.705 &
\textcolor{blue}{0.942} &
0.047 \\

DarkIR \cite{feijoo2025darkir}&
3.320 & 7.250 &
27.684 & 0.875 & 0.115 &
\textcolor{blue}{23.979} &
\textcolor{red}{0.880} &
0.125 &
25.516 & 0.934 & 0.052 \\

CMIG-Net (Ours) &
1.980 & 16.410 &
\textcolor{red}{28.256} &
\textcolor{red}{0.879} &
\textcolor{red}{0.077} &
\textcolor{red}{24.257} &
\textcolor{blue}{0.867} &
\textcolor{blue}{0.122} &
\textcolor{red}{26.324} &
\textcolor{blue}{0.942} &
\textcolor{red}{0.042} \\

\hline
\end{tabular}
}
\captionof{table}{Quantitative comparison on the LOLv1, LOLv2-Real, and
LOLv2-Synthetic datasets. The best and second-best results are
highlighted in red and blue, respectively.}
\label{tab:lol_results}
\end{figure*}

\begin{figure*}[!t]
    \centering
    \setlength{\tabcolsep}{1pt}
    \begin{tabular}{@{}ccccccc@{}}
        &
        \scriptsize 13.81 / 0.824 &
        \scriptsize 16.63 / 0.837 &
        \scriptsize 17.65 / 0.887 &
        \scriptsize 19.94 / 0.885 &
        \scriptsize \textbf{23.67 / 0.910} &
        \\[-2pt]
        \includegraphics[width=0.138\textwidth]
        {\detokenize{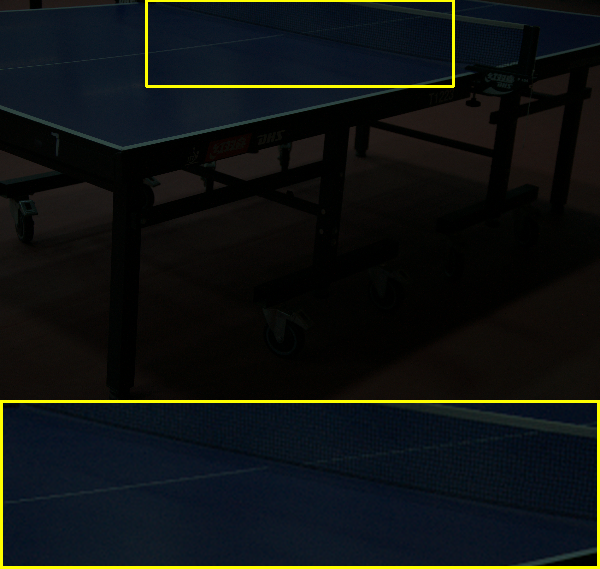}} &
        \includegraphics[width=0.138\textwidth]
        {\detokenize{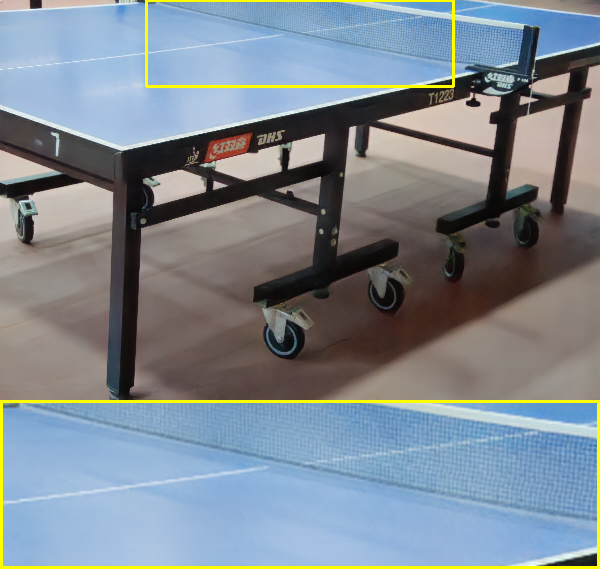}} &
        \includegraphics[width=0.138\textwidth]
        {\detokenize{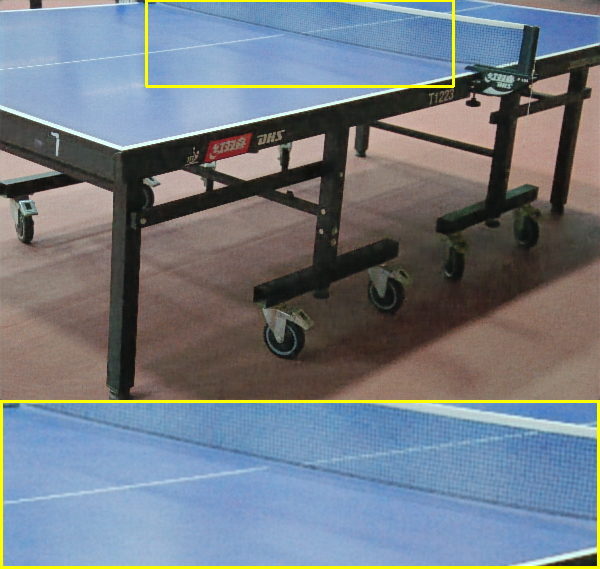}} &
        \includegraphics[width=0.138\textwidth]
        {\detokenize{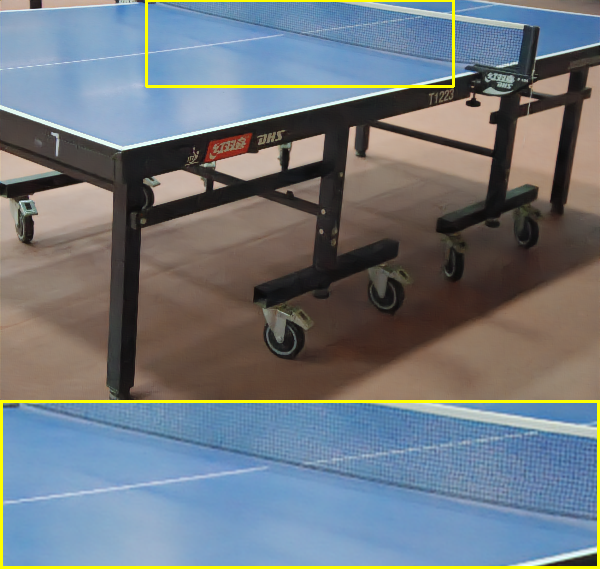}} &
        \includegraphics[width=0.138\textwidth]
        {\detokenize{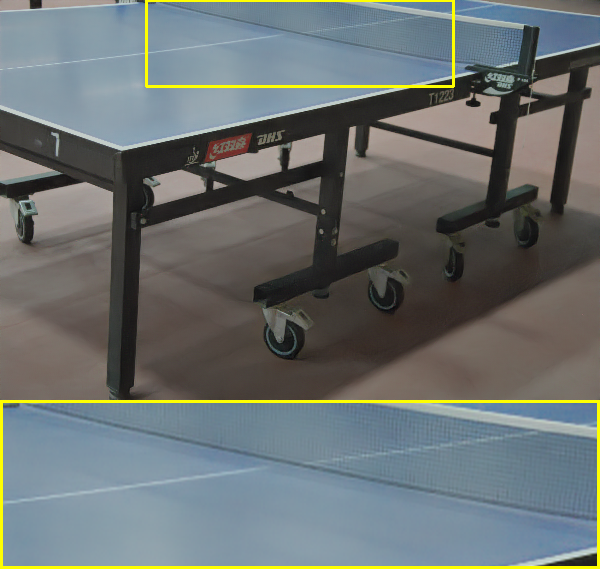}} &
        \includegraphics[width=0.138\textwidth]
        {\detokenize{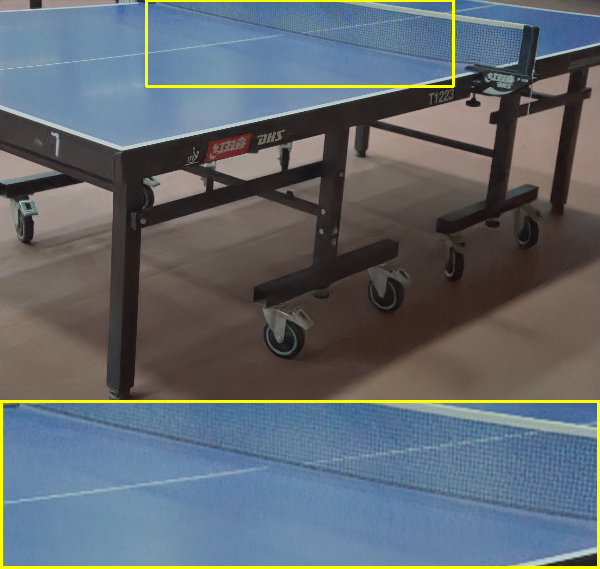}} &
        \includegraphics[width=0.138\textwidth]
        {\detokenize{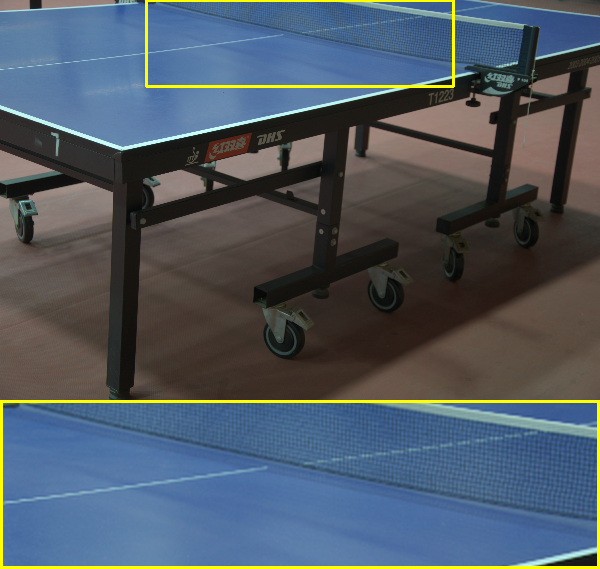}}
        \\[-2pt]

        &
        \scriptsize 28.77 / 0.914 &
        \scriptsize 25.29 / 0.875 &
        \scriptsize 29.86 / 0.903 &
        \scriptsize 28.60 / 0.910 &
        \scriptsize \textbf{31.74 / 0.915} &
        \\[-2pt]
        \includegraphics[width=0.138\textwidth]
        {\detokenize{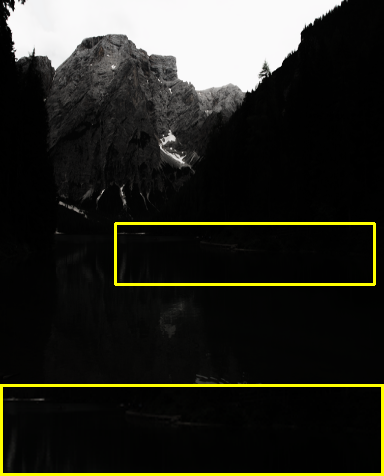}} &
        \includegraphics[width=0.138\textwidth]
        {\detokenize{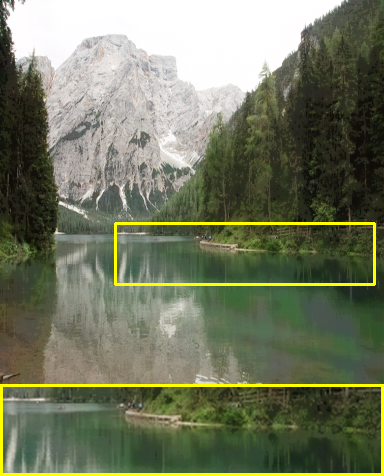}} &
        \includegraphics[width=0.138\textwidth]
        {\detokenize{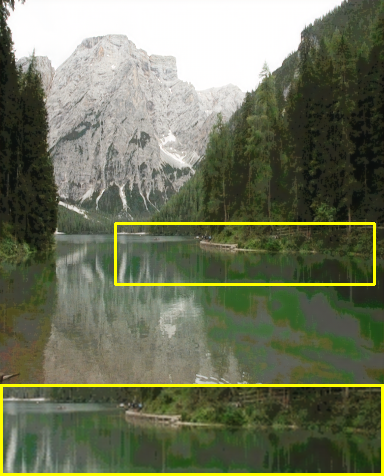}} &
        \includegraphics[width=0.138\textwidth]
        {\detokenize{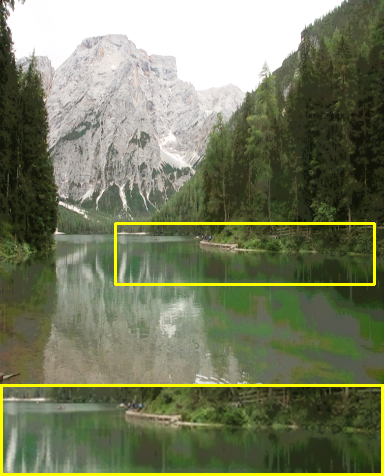}} &
        \includegraphics[width=0.138\textwidth]
        {\detokenize{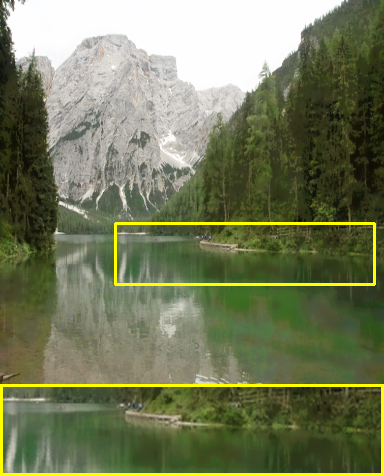}} &
        \includegraphics[width=0.138\textwidth]
        {\detokenize{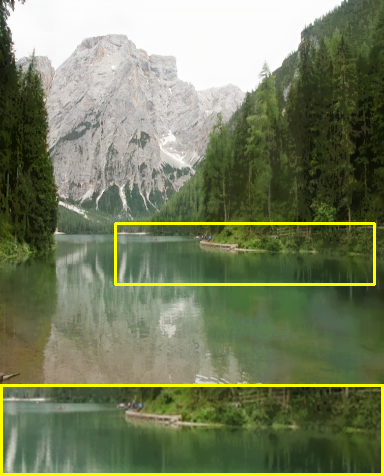}} &
        \includegraphics[width=0.138\textwidth]
        {\detokenize{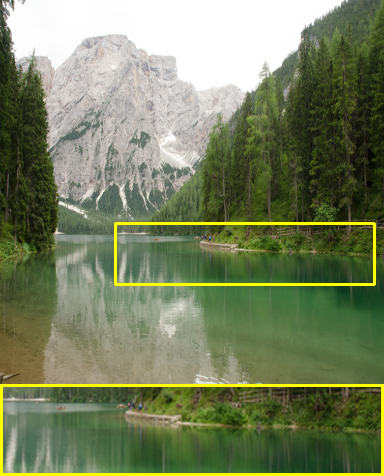}}
        \\[-1pt]

        \footnotesize Input &
        \footnotesize CWNet &
        \footnotesize Retinexformer &
        \footnotesize URWKV &
        \footnotesize CIDNet &
        \footnotesize CMIG-Net &
        \footnotesize Ground Truth
    \end{tabular}
    
    \caption{Qualitative comparison of color and tone restoration by different LLIE methods on the LOL dataset. The numbers above each result denote PSNR (dB) / SSIM.}
    \label{fig:03}
    
\end{figure*}

\subsection{Dynamic Dual-branch Information Restoration Module}\label{sec:D2IR}

Existing methods perform dual-branch restoration according to the distinct information properties and degradation patterns of $I$ and $HV$. They also model information exchange between the two components \cite{yan2025hvi, xu2026iclr}. Nevertheless, the useful contribution of $HV$ given $I$ varies across spatial locations and restoration stages. Fixed cross-branch interactions cannot accommodate this dynamic demand.
To address this limitation, we propose the Dynamic Dual-branch Information Restoration (D2IR) module, which introduces a dynamic interaction mechanism guided by conditional information. D2IR treats $M_{\mathrm{cmi}}$ as a conditional information prior and combines it with the current dual-branch feature states. The resulting bidirectional updates are formulated as:
\begin{equation}
\begin{array}{c}
F_I^{t+1}
=
\mathrm{CIGA}_I^t
\left(F_I^t,F_{HV}^t,M_{\mathrm{cmi}}\right), \\[2pt]
F_{HV}^{t+1}
=
\mathrm{CIGA}_{HV}^t
\left(F_{HV}^t,F_I^t,M_{\mathrm{cmi}}\right).
\end{array}
\label{eq:09}
\end{equation}
In Eq. \ref{eq:09}, $F_I^t$ and $F_{HV}^t$ denote the $I$ and $HV$ features, respectively, at the $t$-th restoration stage. The first input to CIGA is the target-branch feature to be updated. The second is the source-branch feature that provides complementary information. $\mathrm{CIGA}_I^t$ updates the $I$ branch under the guidance of the $HV$ branch, whereas $\mathrm{CIGA}_{HV}^t$ updates the $HV$ branch under the guidance of the $I$ branch. This process preserves the structural advantages of complementary I/HV dual-branch restoration. It also extends generic cross-branch fusion to dynamic restoration driven by conditional information.

\begin{figure*}[t]
    \centering
    \setlength{\tabcolsep}{1pt}
    \begin{tabular}{@{}ccccccc@{}}
        \includegraphics[width=0.138\textwidth]
        {\detokenize{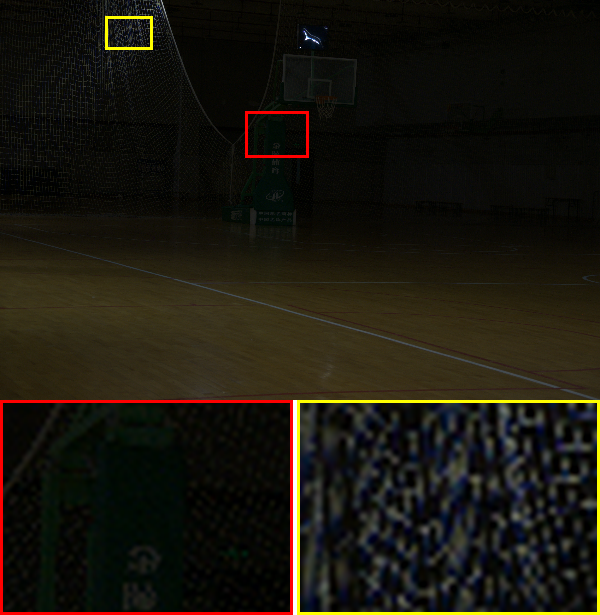}} &
        \includegraphics[width=0.138\textwidth]
        {\detokenize{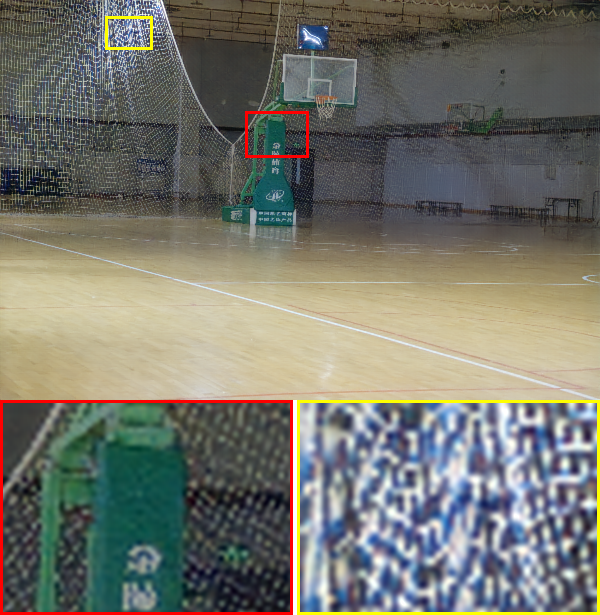}} &
        \includegraphics[width=0.138\textwidth]
        {\detokenize{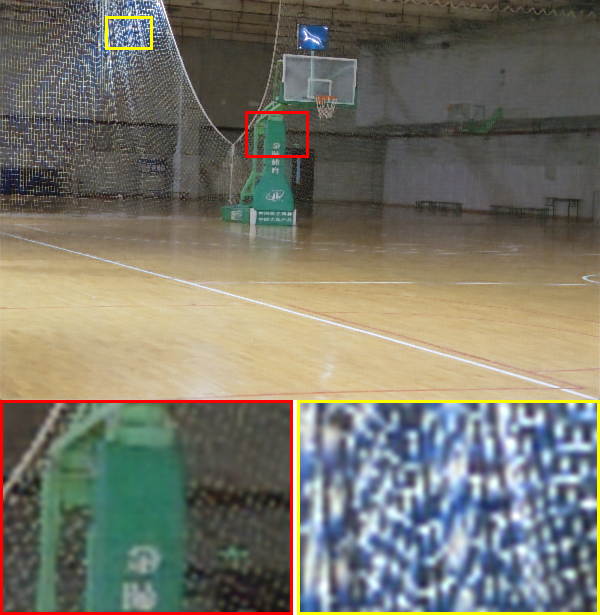}} &
        \includegraphics[width=0.138\textwidth]
        {\detokenize{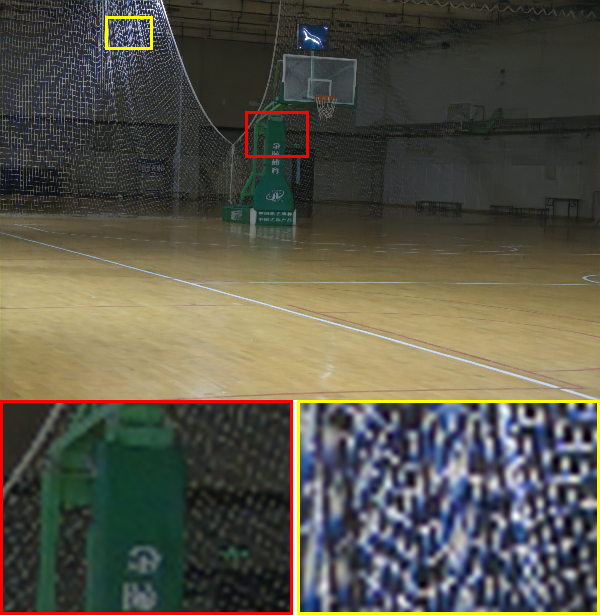}} &
        \includegraphics[width=0.138\textwidth]
        {\detokenize{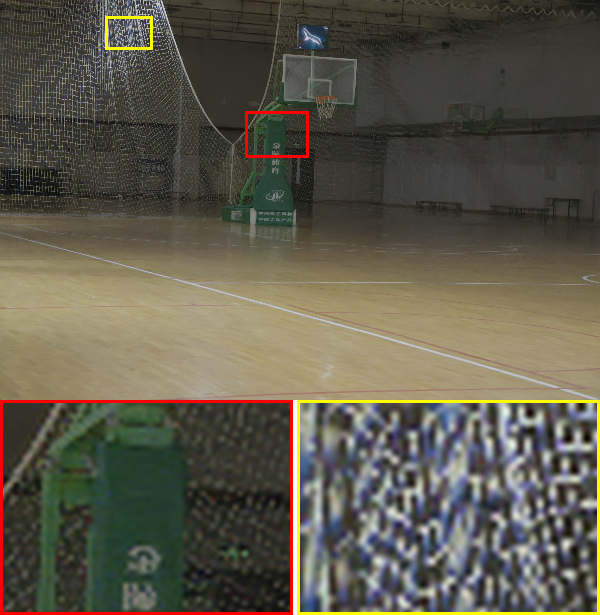}} &
        \includegraphics[width=0.138\textwidth]
        {\detokenize{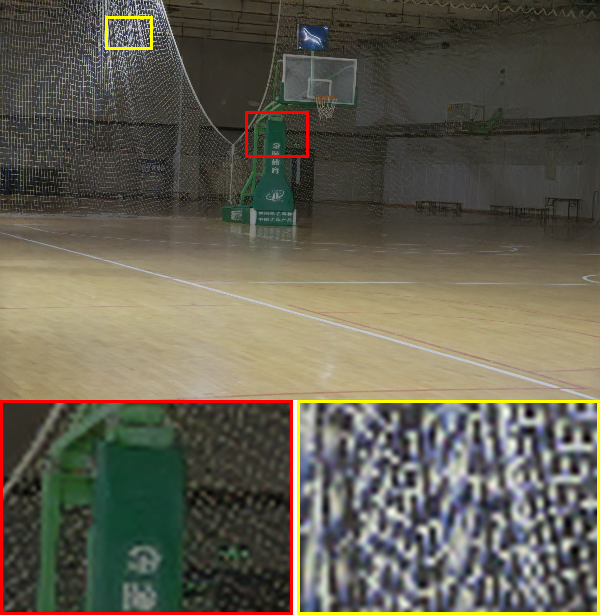}} &
        \includegraphics[width=0.138\textwidth]
        {\detokenize{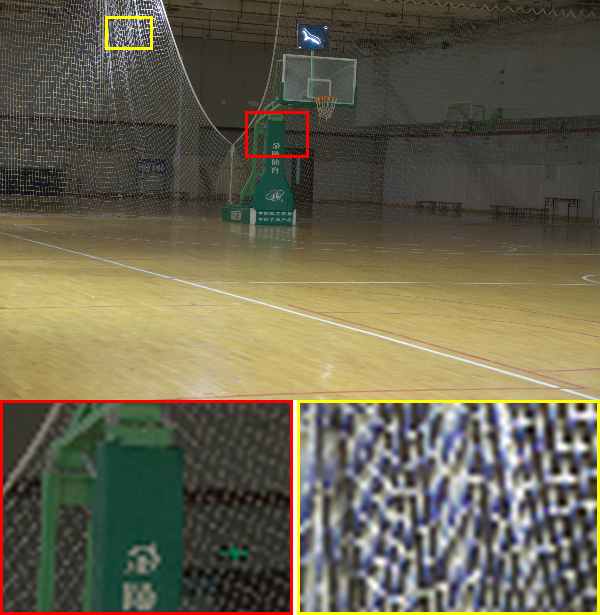}}
        \\[-1pt]

        \includegraphics[width=0.138\textwidth]
        {\detokenize{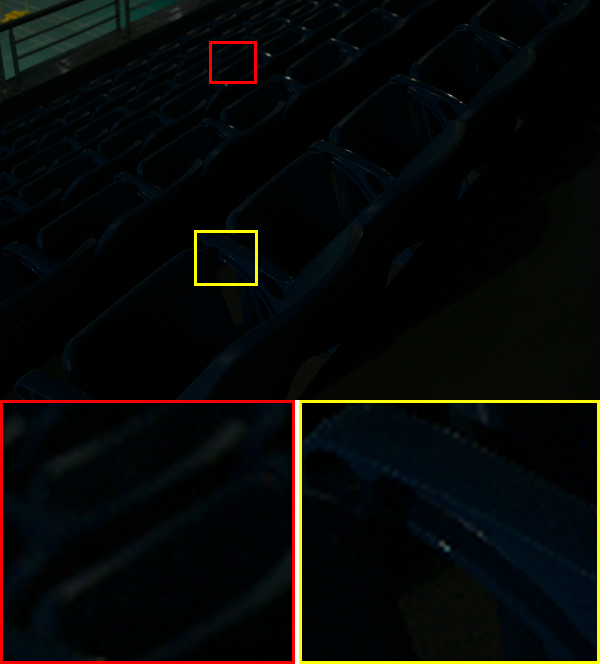}} &
        \includegraphics[width=0.138\textwidth]
        {\detokenize{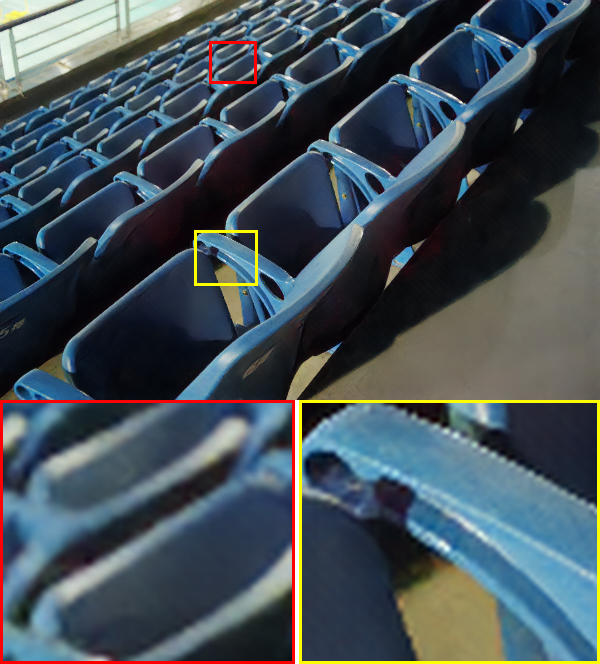}} &
        \includegraphics[width=0.138\textwidth]
        {\detokenize{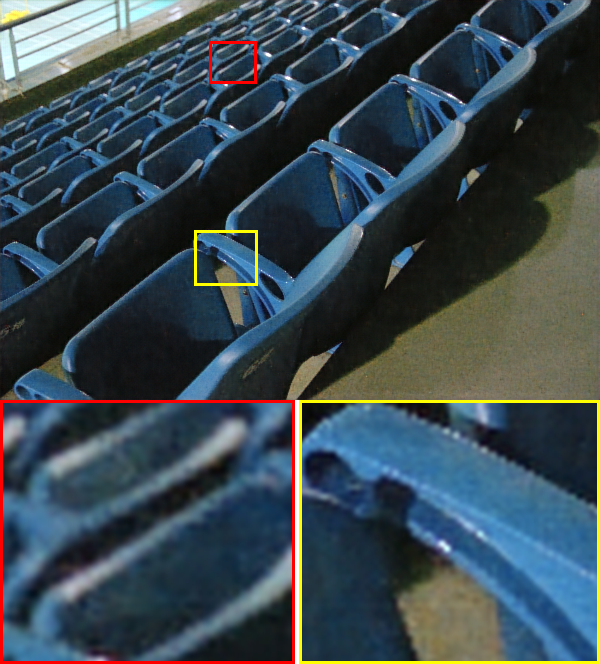}} &
        \includegraphics[width=0.138\textwidth]
        {\detokenize{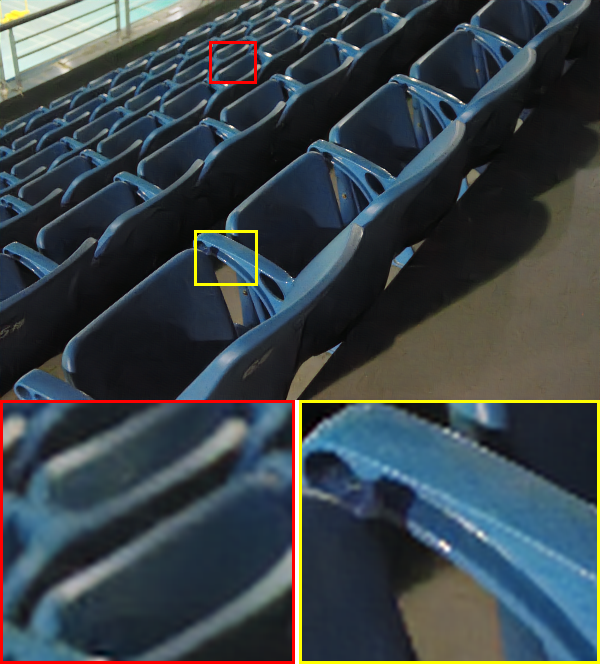}} &
        \includegraphics[width=0.138\textwidth]
        {\detokenize{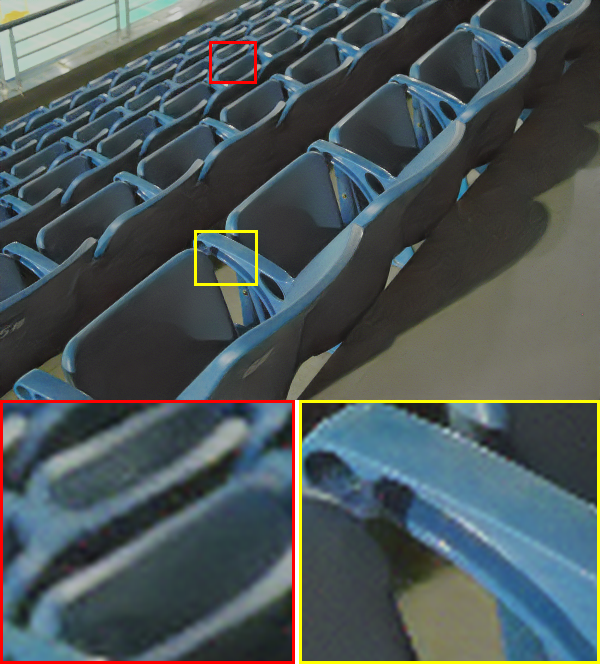}} &
        \includegraphics[width=0.138\textwidth]
        {\detokenize{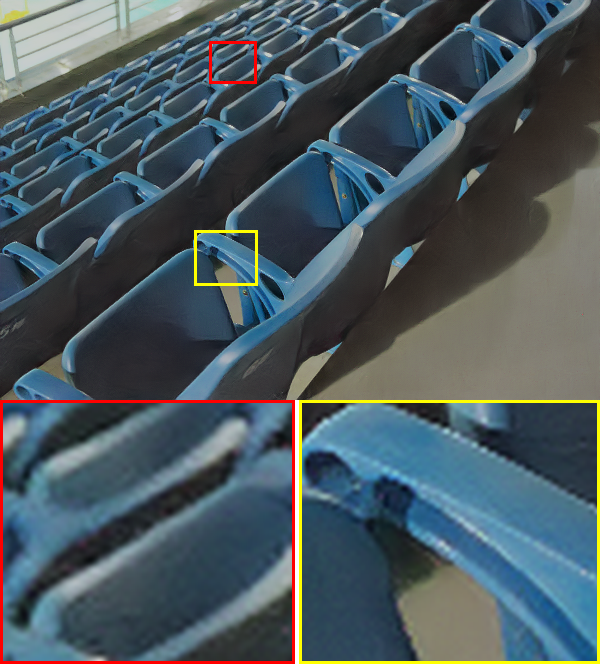}} &
        \includegraphics[width=0.138\textwidth]
        {\detokenize{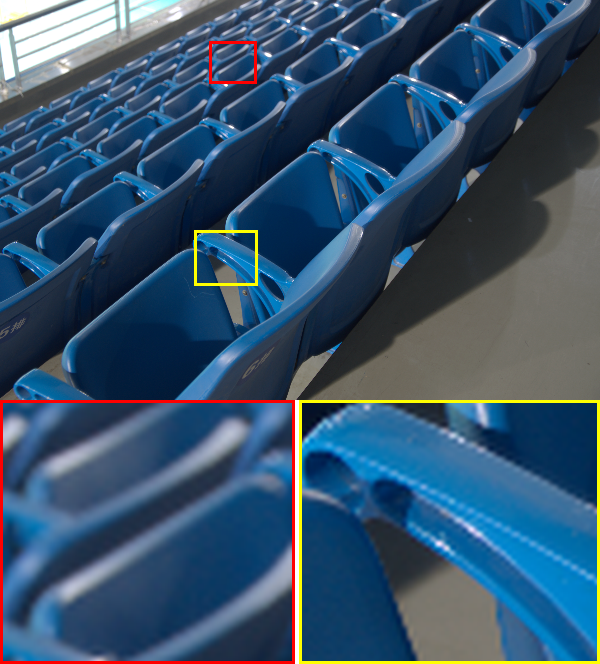}}
        \\[-1pt]

        \footnotesize Input &
        \footnotesize CWNet &
        \footnotesize Retinexformer &
        \footnotesize URWKV &
        \footnotesize CIDNet &
        \footnotesize CMIG-Net (Ours) &
        \footnotesize Ground Truth
    \end{tabular}
   
    \caption{Qualitative comparison of detail recovery and noise suppression by different LLIE methods on the LOL dataset.}
    \label{fig:04}
\end{figure*}

\subsection{Conditional Information-Guided Attention (CIGA)}\label{sec:CIGA}

The $M_{\mathrm{cmi}}$ provided by CMIC is a static input-level prior, whereas the demand for cross-branch information varies across restoration stages. CIGA therefore combines the conditional information prior with the current dual-branch states to dynamically regulate information exchange. At stage $t$, let $X$ denote the target branch to be updated and $Y$ denote the source branch that provides complementary information. First, CIGA constructs the state-aware guidance as:
\begin{equation}
G^t
=
\mathrm{Concat}
\left(
M_{\mathrm{cmi}}^t,
\mathrm{Avg}_c(X),
\mathrm{Avg}_c(Y)
\right).
\label{eq:10}
\end{equation}

In Eq. \ref{eq:10}, $M_{\mathrm{cmi}}^t$ is the conditional information map resized to the current feature resolution, and $\mathrm{Avg}_c(\cdot)$ denotes channel-wise averaging. The matching relationship between the target and source branches is established through $Q_x$, $K_y$, and $V_y$.
CIGA applies state-aware modulation before and after information aggregation. First, the value representation of the source branch is modulated as
\begin{equation}
\begin{array}{c}
g_v^t
=
1+\gamma\tanh\!\left(f_v(G^t)\right), 
\widehat{V}_y
=
V_y\odot g_v^t.
\end{array}
\label{eq:11}
\end{equation}
Using the modulated value in Eq. \ref{eq:11}, cross-branch attention aggregation is performed as transposed channel attention:
\begin{equation}
\begin{array}{c}
A
=
\mathrm{Softmax}
\left(
\widehat{Q}_x
\widehat{K}_y^{\top}
\odot \tau
\right), 
O
=
A\widehat{V}_y.
\end{array}
\label{eq:12}
\end{equation}
In Eq. \ref{eq:12}, $\widehat{Q}_x$ and $\widehat{K}_y$ are $\ell_2$-normalized features and $\tau$ is a learnable temperature parameter. Softmax is applied along the channel dimension of the affinity matrix. Finally, the aggregated result is modulated at the output and injected into the target branch as:
\begin{equation}
\begin{array}{c}
g_o^t
=
1+\gamma\tanh\!\left(f_o(G^t)\right), 
X'
=
X+W_0\!\left(O\odot g_o^t\right).
\end{array}
\label{eq:13}
\end{equation}
Together, Eq. \ref{eq:11} and Eq. \ref{eq:13} show that value modulation controls the source information used in aggregation, whereas output modulation controls its final injection strength. CIGA therefore performs adaptive cross-branch information exchange according to the conditional contribution and current restoration state.

\subsection{Training Objectives}\label{sec:loss}

Without explicit supervision, the reference distribution $P$ predicted solely from the intensity input $I$ tends to collapse. The resulting CMI estimate then becomes physically meaningless.
We introduce a Reference Distribution Fitting Loss, $L_{\mathrm{p-fit}}$. This loss uses the full-information distribution $Q$ learned by the network as the target and aligns the reference distribution $P$ with $Q$. Consequently, $P$ faithfully captures the cluster states that can be inferred from intensity alone when chrominance information is unavailable. The loss is defined as

\begin{equation}
L_{\mathrm{p-fit}}
=
\frac{1}{N}
\sum_{i=1}^{N}
\sum_{c=1}^{K}
Q_c
\log\left(
\frac{Q_c+\epsilon}
     {P_c+\epsilon}
\right).
\label{eq:14}
\end{equation}

In Eq. \ref{eq:14}, $N = B \times H \times W$ denotes the total number of pixels in a batch. This loss is used exclusively to optimize the branch that generates $P$. It provides a stable and accurate intensity-only reference for computing CMI.
During training, we jointly optimize the base reconstruction loss $L_{rec}$, which constrains restoration quality, and the key regularization term $L_{\mathrm{p-fit}}$. The total training objective is defined as

\begin{equation}
L_{\mathrm{total}}
=
L_{\mathrm{rec}}
+
\lambda_{\mathrm{fit}}L_{\mathrm{p-fit}}.
\label{eq:15}
\end{equation}
The objective in Eq. \ref{eq:15} enables the network to extract conditional information reliably and use it to guide cross-branch feature injection (Supplementary Material, Sec.~B).

\section{Experiments}\label{sec:Experiment}

\subsection{Datasets and Settings}\label{sec:Datasets and Settings}

\noindent\textbf{Datasets.}
We conduct paired evaluations on LOLv1 \cite{DBLP:conf/bmvc/WeiWY018}, LOLv2-Real and LOLv2-Synthetic \cite{yang2021sparse}, SICE \cite{cai2018learning, DBLP:journals/corr/abs-2212-10772}, and Sony-Total-Dark (SID) \cite{chen2018learning}. We evaluate generalization to unpaired settings on DICM \cite{lee2013contrast}, LIME \cite{guo2016lime}, MEF \cite{ma2015perceptual}, NPE \cite{wang2013naturalness}, and VV \cite{vonikakis2018evaluation}. The raw SID data are converted to sRGB images without gamma correction.

\noindent\textbf{Training Settings.}
For LOLv1 and LOLv2-Real, training images are randomly cropped into $400 \times 400$ patches. The models are trained for 1,500 epochs with a batch size of 8. For LOLv2-Synthetic, the model is trained without cropping for 500 epochs with a batch size of 1. For SICE and SID, we use $400 \times 400$ and $512 \times 512$ patches, respectively. Both models are trained for 500 epochs, by which point the optimum has already been reached, with batch sizes of 8 and 4, respectively. The model is implemented in PyTorch and trained on a single NVIDIA V100 GPU. Adam \cite{DBLP:journals/corr/KingmaB14} is used with an initial learning rate of $1\times10^{-4}$, which is reduced to $1\times10^{-7}$ by cosine annealing \cite{loshchilov2017sgdr}.

\noindent\textbf{Evaluation.}
We use PSNR, SSIM \cite{wang2004image}, and LPIPS \cite{zhang2018unreasonable} for paired datasets, and NIQE \cite{mittal2012making} and BRISQUE \cite{mittal2012no} for unpaired datasets. All ablation variants are trained and evaluated using the same settings. For competing methods, we use their official implementations or publicly reported results (Supplementary Material, Sec.~C).

\subsection{Main Results}\label{sec:Main Results}

\noindent\textbf{Results on LOL Datasets.}
As shown in Table \ref{tab:lol_results}, CMIG-Net achieves the best PSNR on all three LOL benchmarks, with 28.256 dB on LOLv1, 24.257 dB on LOLv2-Real, and 26.324 dB on LOLv2-Synthetic. It also obtains the lowest LPIPS on LOLv1 and LOLv2-Synthetic while maintaining competitive SSIM. On LOLv1, it outperforms CIDNet, improving PSNR from 27.890 to 28.256 dB and LPIPS from 0.080 to 0.077.

The visual comparisons in Figure \ref{fig:03} show more faithful colors and smoother transitions in the table-tennis table and lake scenes. In Figure \ref{fig:04}, CMIG-Net restores clearer court and seat structures with less noise and fewer chrominance artifacts.

\noindent\textbf{Results on Additional Paired and Unpaired Benchmarks.} As shown in Table \ref{tab:sice_sid_unpaired}, CMIG-Net achieves the highest SSIM on SICE and the best PSNR and SSIM on SID. On the latter, it outperforms CIDNet by 0.382 dB and 0.011 in PSNR and SSIM, respectively. Across five unpaired datasets, CMIG-Net obtains the best BRISQUE score of 20.458 and the second-best NIQE score of 3.545.
Figure \ref{fig:unpaired_visual_comparison} further shows improved visibility and more natural colors across diverse real-world scenes, with less underexposure, over-enhancement, and color shift. These results confirm robust generalization to both paired and unpaired benchmarks (Supplementary Material, Secs.~F and~G).

\begin{table}[!tb]
\centering
\scriptsize
\setlength{\tabcolsep}{3pt}
\renewcommand{\arraystretch}{0.8}
\resizebox{\linewidth}{!}{
\begin{tabular}{lcccccc}
\toprule
\textbf{Method}
& \multicolumn{2}{c}{\textbf{SICE}}
& \multicolumn{2}{c}{\textbf{SID}}
& \multicolumn{2}{c}{\textbf{Unpaired}} \\
& PSNR$\uparrow$
& SSIM$\uparrow$
& PSNR$\uparrow$
& SSIM$\uparrow$
& BRISQUE$\downarrow$
& NIQE$\downarrow$ \\
\midrule
RetinexNet 
& 12.424 & 0.613
& 15.695 & 0.395
& 23.286 & 4.558 \\
Zero-DCE 
& 12.452 & 0.639
& 14.087 & 0.090
& 26.343 & 4.763 \\
URetinexNet 
& 10.899 & 0.605
& 15.519 & 0.323
& 26.359 & 3.829 \\
RUAS 
& 8.656 & 0.494
& 12.622 & 0.081
& 36.372 & 4.800 \\
LLFlow 
& 12.737 & 0.617
& 16.226 & 0.367
& 28.087 & 4.221 \\
CIDNet
& \textbf{13.435} & 0.642
& 22.904 & 0.676
& 23.521 & \textbf{3.523} \\
\textbf{CMIG-Net (Ours)}
& 13.207 & \textbf{0.661}
& \textbf{23.286} & \textbf{0.687}
& \textbf{20.458} & 3.545 \\
\bottomrule
\end{tabular}
}
\caption{Quantitative comparison on SICE, SID, and unpaired datasets. The best results are shown in bold.}
\label{tab:sice_sid_unpaired}
\end{table}

\begin{figure}[!tb]
    \centering
    \setlength{\tabcolsep}{1pt}

    \newcommand{\cmpimg}[2]{%
        \shortstack{%
            \includegraphics[width=0.235\columnwidth]
            {Figures/Unpaired_compare/#2}\\[-2pt]
            \scriptsize #1
        }%
    }

    \begin{tabular}{@{}cccc@{}}

        \cmpimg{Input}{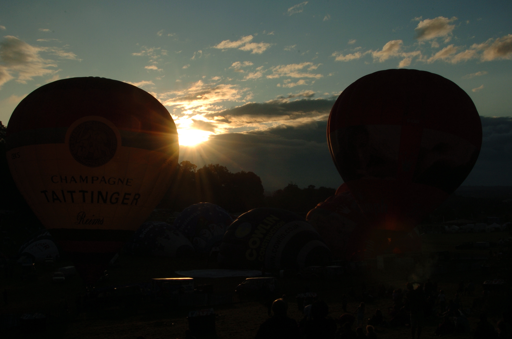} &
        \cmpimg{URetinexNet}{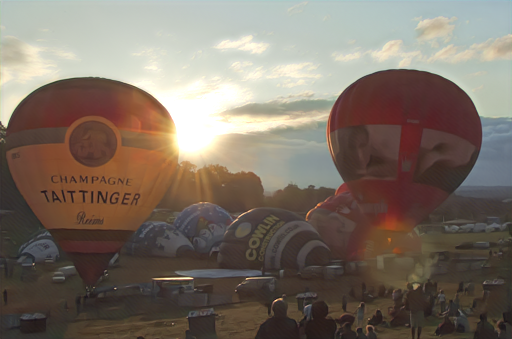} &
        \cmpimg{CIDNet}{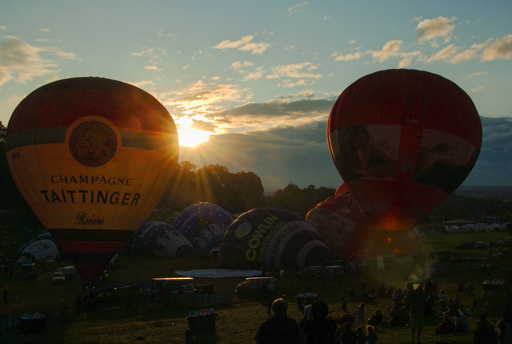} &
        \cmpimg{\textbf{Ours}}{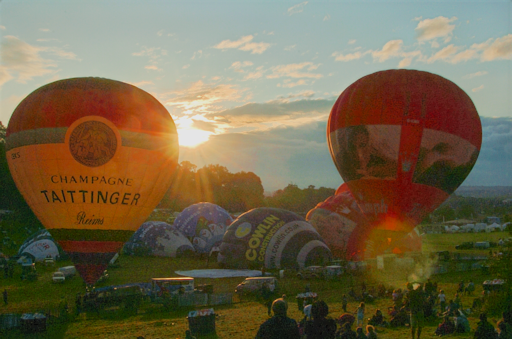}
        \\[-2pt]

        \cmpimg{Input}{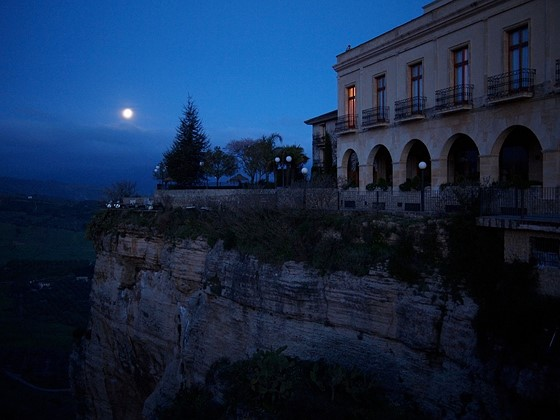} &
        \cmpimg{Zero-DCE}{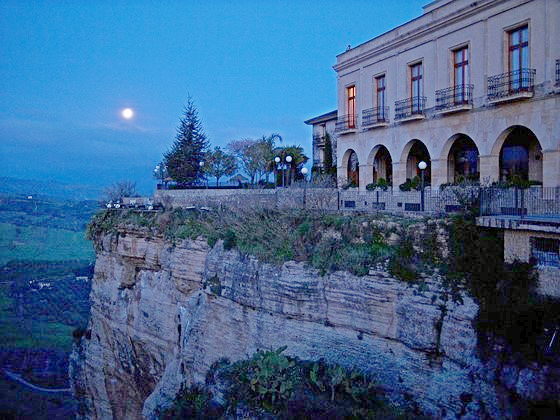} &
        \cmpimg{CIDNet}{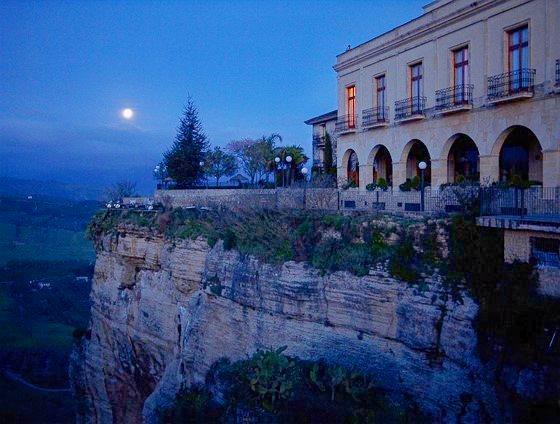} &
        \cmpimg{\textbf{Ours}}{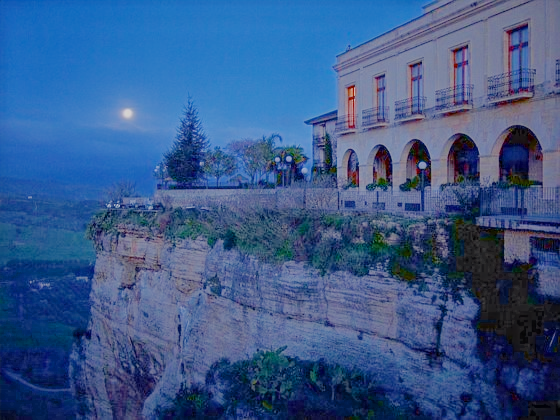}
        \\[-2pt]

        \cmpimg{Input}{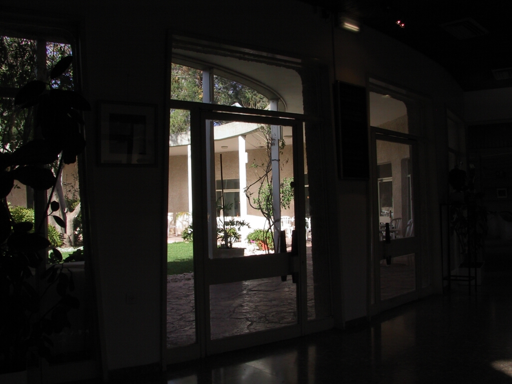} &
        \cmpimg{Retinexformer}{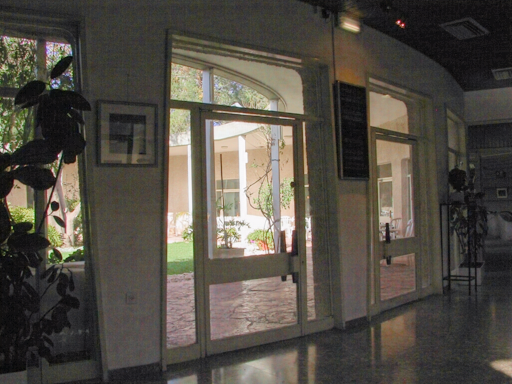} &
        \cmpimg{Zero-DCE}{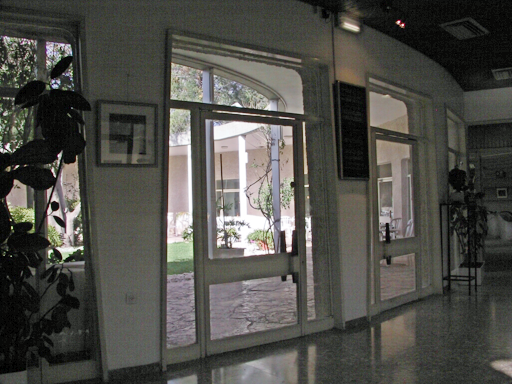} &
        \cmpimg{\textbf{Ours}}{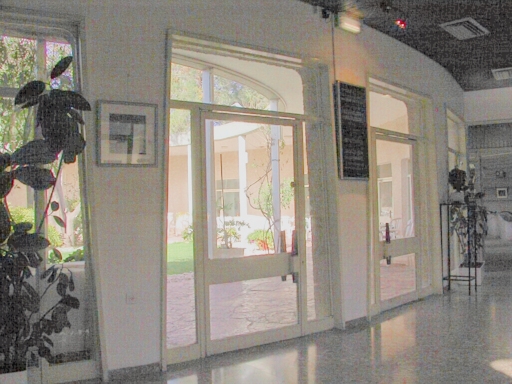}
        \\[-2pt]

        \cmpimg{Input}{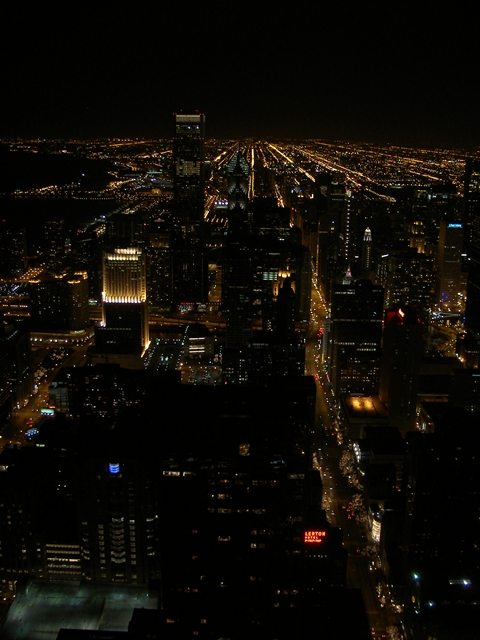} &
        \cmpimg{LLFlow}{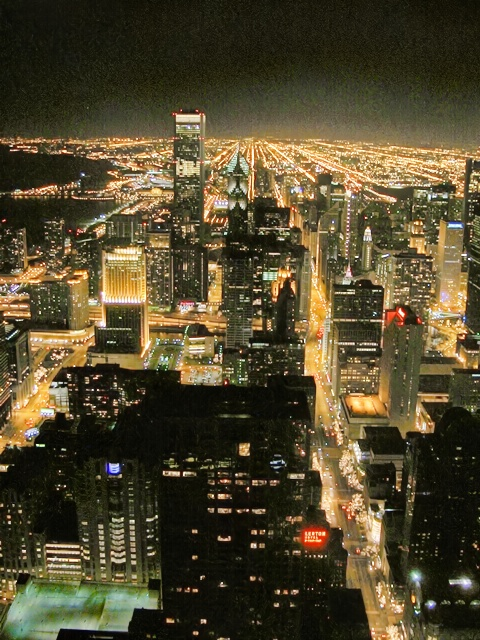} &
        \cmpimg{Retinexformer}{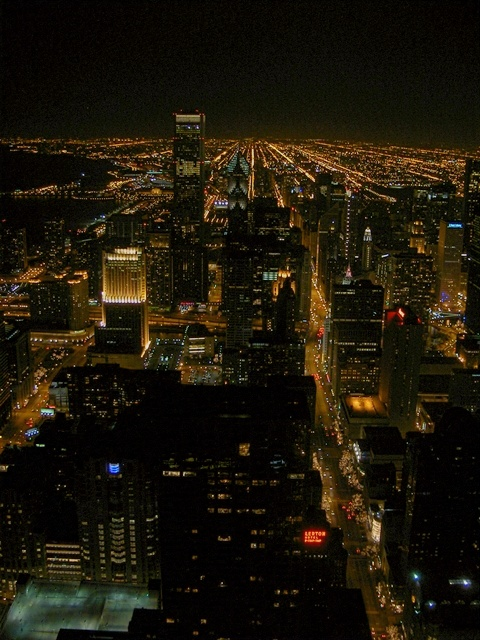} &
        \cmpimg{\textbf{Ours}}{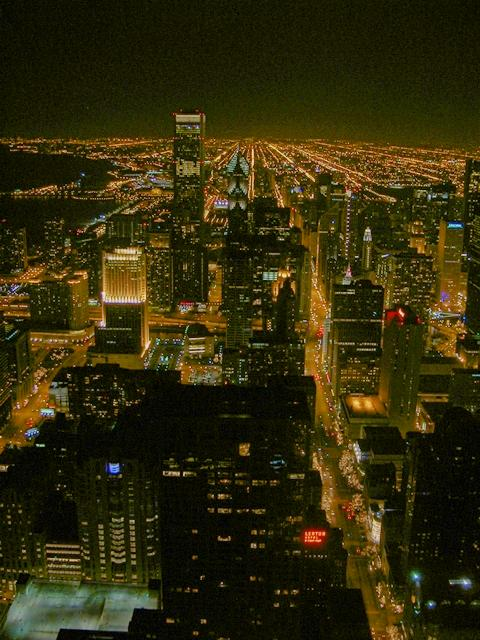}

    \end{tabular}

    \caption{Qualitative comparisons on unpaired low-light datasets.}
    \label{fig:unpaired_visual_comparison}
\end{figure}

\subsection{Ablation Study}\label{sec:Ablation Study}

To systematically evaluate each design in CMIG-Net, we conduct quantitative ablations and visual analyses at three levels: overall components, CMIC conditional calibration, and D2IR dynamic information injection. All experiments use consistent training and evaluation settings on LOLv1 or LOLv2 to ensure fair comparisons (Supplementary Material, Secs.~D and~E).

\begin{figure}[!tb]
    \centering
    \setlength{\tabcolsep}{1.2pt}
    \newcommand{\ablpanel}[2]{%
        \begin{minipage}[t]{0.158\linewidth}
            \centering
            \includegraphics[width=\linewidth]{#1}\\
            \scriptsize #2
        \end{minipage}%
    }
    \ablpanel{\detokenize{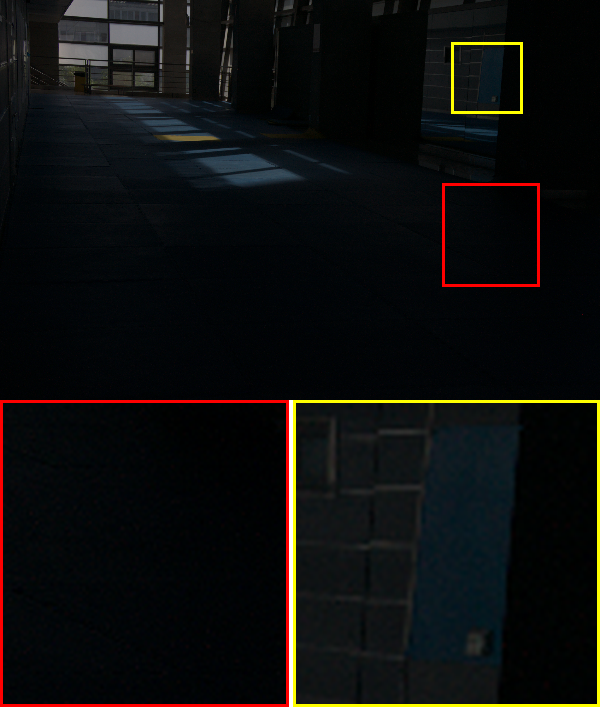}}{Input}\hfill
    \ablpanel{\detokenize{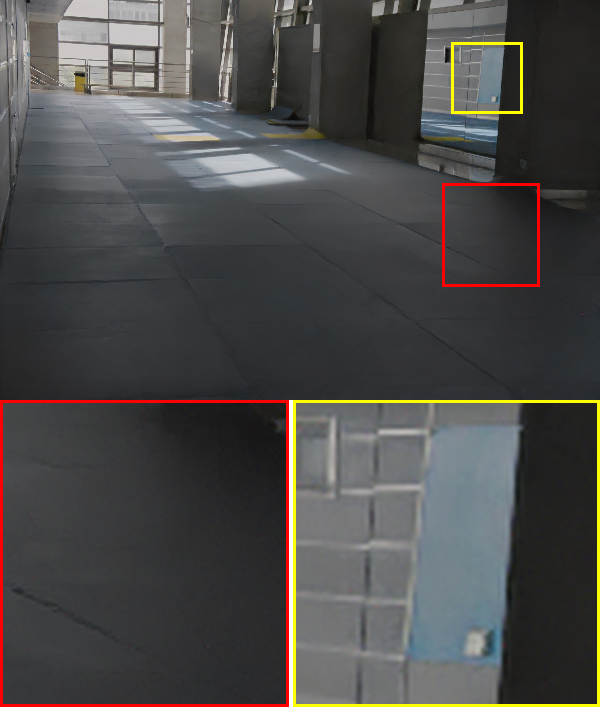}}{Baseline}\hfill
    \ablpanel{\detokenize{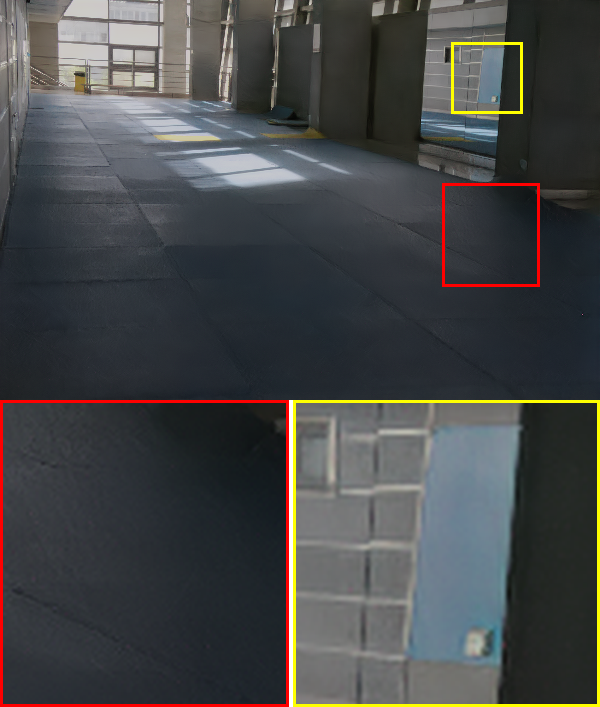}}{CMI}\hfill
    \ablpanel{\detokenize{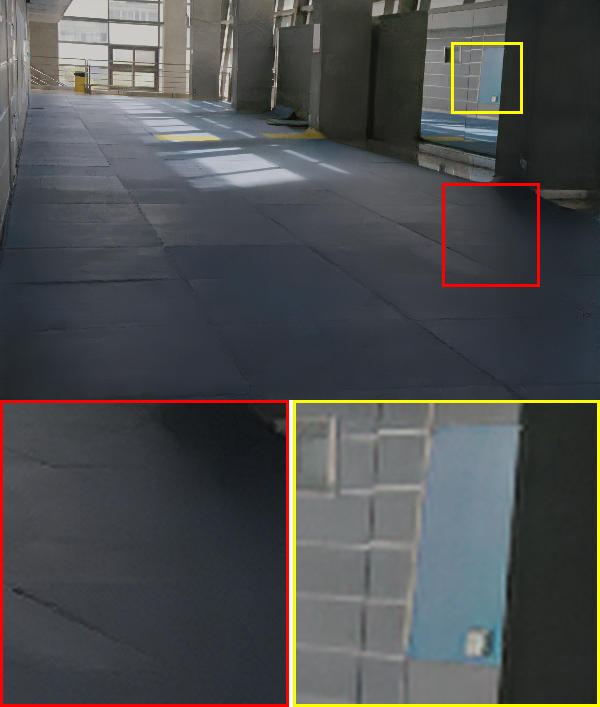}}{Static CIGA}\hfill
    \ablpanel{\detokenize{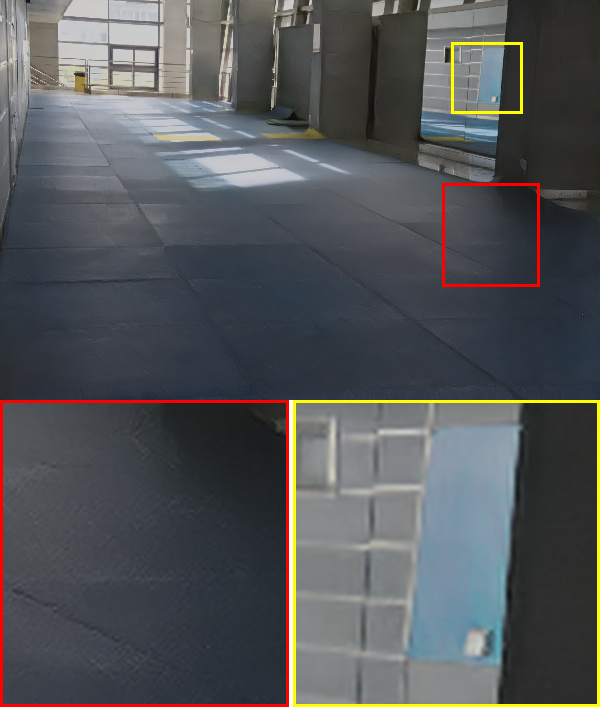}}{Dynamic CIGA}\hfill
    \ablpanel{\detokenize{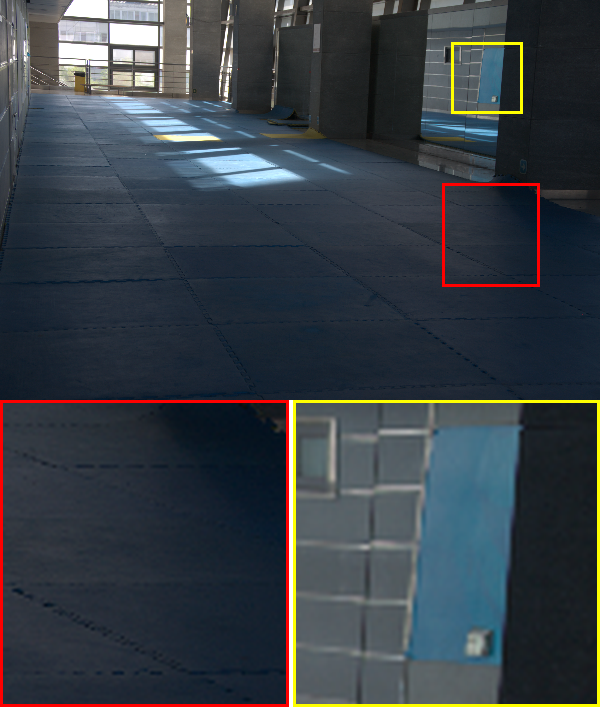}}{Ground Truth}\\[2pt]
    \ablpanel{\detokenize{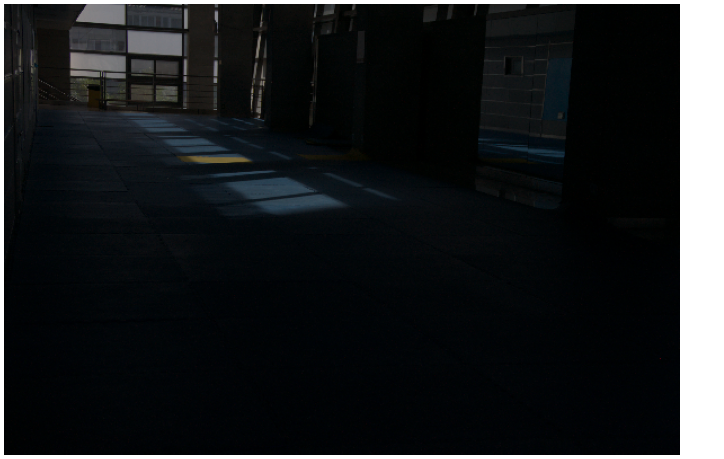}}{Input}\hfill
    \ablpanel{\detokenize{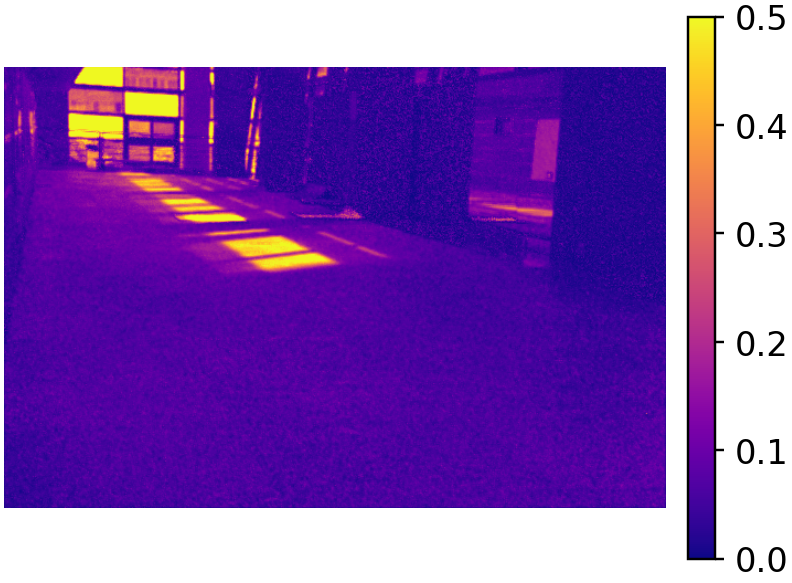}}{$M_{\mathrm{cmi}}$}\hfill
    \ablpanel{\detokenize{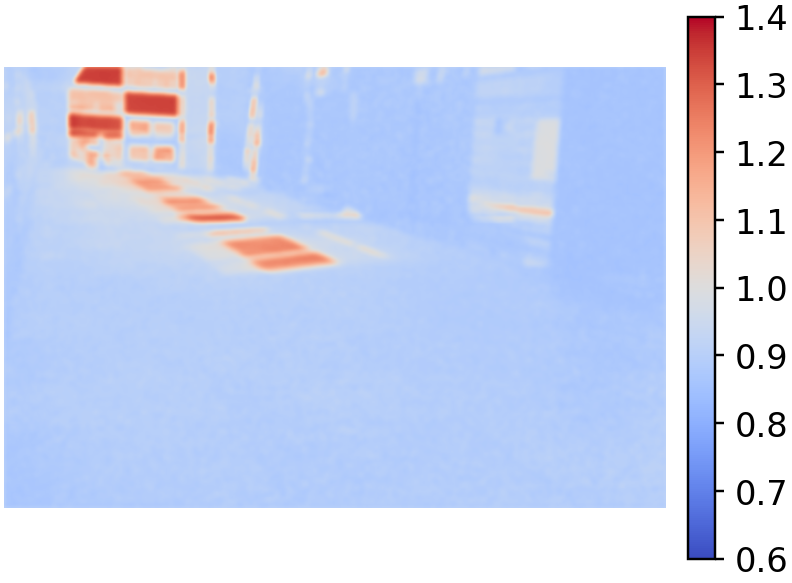}}{Early gate}\hfill
    \ablpanel{\detokenize{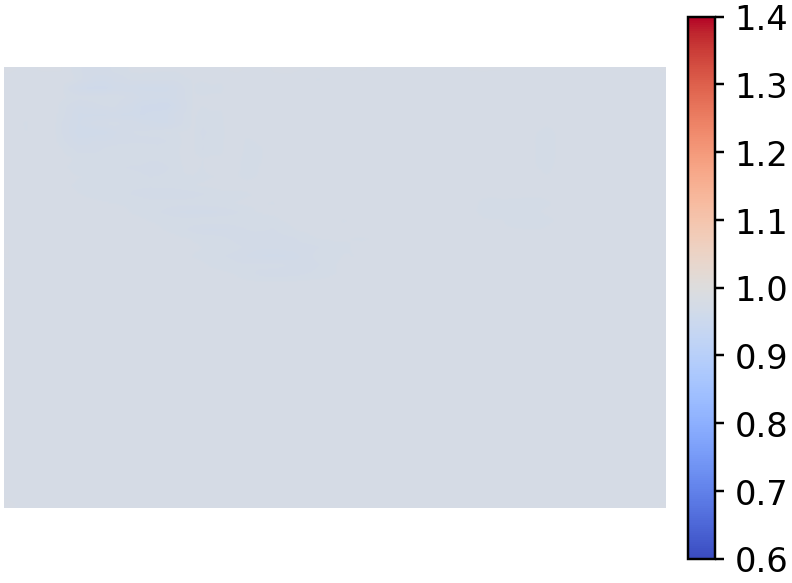}}{Bottleneck}\hfill
    \ablpanel{\detokenize{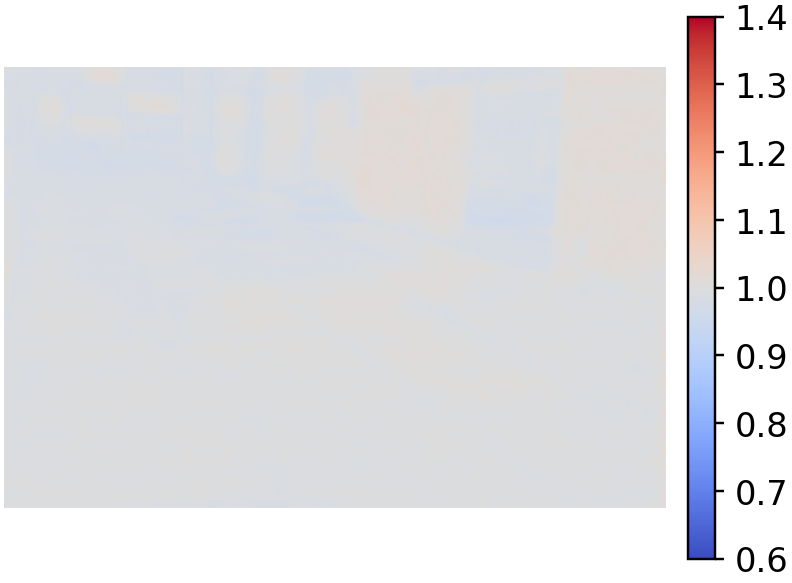}}{Late gate}\hfill
    \ablpanel{\detokenize{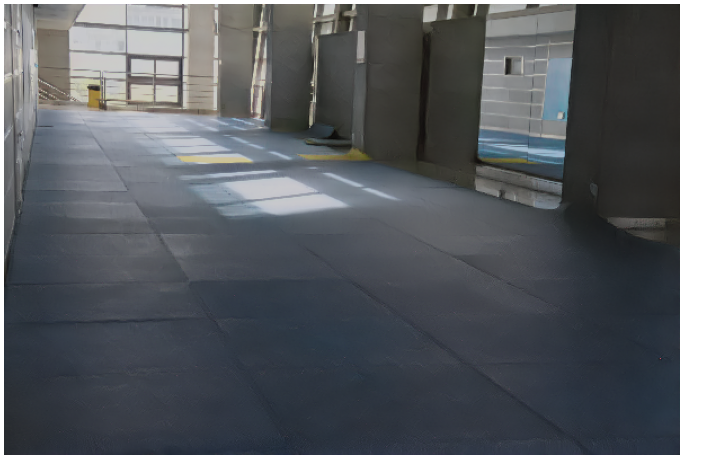}}{Output}
    \caption{Visual analysis of conditional information calibration and
    dynamic cross-branch information injection, including module configurations,
    the spatial CMI prior, and stage-wise $HV$-to-$I$ injection gates.}
    \label{fig:ablation_cmi_visualization}
\end{figure}

\noindent\textbf{Component-wise Analysis of CMIG-Net.} As shown in Table \ref{tab:component_ablation}, CMIC improves the baseline PSNR from 23.331 to 23.672 dB, whereas Static CIGA achieves only 23.405 dB. Dynamic CIGA reaches 24.126 dB, outperforming Static CIGA and the baseline by 0.721 and 0.795 dB, respectively, with the best SSIM and LPIPS. Figure \ref{fig:ablation_cmi_visualization} further shows improved color recovery and noise suppression, while the spatially varying $M_{\mathrm{cmi}}$ and stage-dependent gates demonstrate adaptive $HV$ injection.

\begin{table}[!tb]
    \centering
    \small
    \setlength{\tabcolsep}{7pt}
    \renewcommand{\arraystretch}{0.8}

    \resizebox{\columnwidth}{!}{
        \begin{tabular}{ccc|ccc}
            \toprule
            \multirow{2}{*}{\textbf{CMIC}}
            & \multicolumn{2}{c|}{\textbf{CIGA}}
            & \multirow{2}{*}{PSNR$\uparrow$}
            & \multirow{2}{*}{SSIM$\uparrow$}
            & \multirow{2}{*}{LPIPS$\downarrow$} \\
            \cmidrule(lr){2-3}
            & \textbf{Static} & \textbf{Dynamic} & & & \\
            \midrule
            & & & 23.331 & 0.860 & 0.123 \\
            $\checkmark$ & & & 23.672 & 0.865 & 0.122 \\
            $\checkmark$ & $\checkmark$ & & 23.405 & 0.863 & 0.119 \\
            $\checkmark$ & & $\checkmark$ & 24.126 & 0.868 & 0.113 \\
            \bottomrule
        \end{tabular}
    }
    \caption{Component-wise ablation study of CMIG-Net on LOLv2-Real.}
    \label{tab:component_ablation}
\end{table}

\noindent\textbf{Effectiveness of CMIC.} As shown in Table \ref{tab:cmic_ablation}, CMI Modeling improves PSNR by 0.382 dB. Adding HV Calibration achieves the best SSIM of 0.880 and LPIPS of 0.076. With $L_{\mathrm{p-fit}}$, PSNR further increases to 28.256 dB, 0.468 dB above the baseline, while SSIM and LPIPS slightly change to 0.879 and 0.077, respectively. 

\begin{table}[!tb]
    \centering
    \small
    \setlength{\tabcolsep}{4pt}
    \renewcommand{\arraystretch}{0.8}

    \resizebox{\columnwidth}{!}{
        \begin{tabular}{cccccc}
            \toprule
            \textbf{CMI Modeling}
            & \textbf{HV Calibration}
            & \textbf{$L_{\mathrm{p-fit}}$}
            & PSNR$\uparrow$
            & SSIM$\uparrow$
            & LPIPS$\downarrow$ \\
            \midrule
            & & & 27.788 & 0.879 & 0.078 \\
            $\checkmark$ & & & 28.170 & 0.879 & 0.080 \\
            $\checkmark$ & $\checkmark$ & & 28.060 & 0.880 & 0.076 \\
            $\checkmark$ & $\checkmark$ & $\checkmark$
            & 28.256 & 0.879 & 0.077 \\
            \bottomrule
        \end{tabular}
    }
    \caption{Ablation study of the individual components in CMIC on LOLv1.}
    \label{tab:cmic_ablation}
\end{table}

\noindent\textbf{Effectiveness of D2IR.}
As shown in Table \ref{tab:d2ir_ablation}, Static guidance on the source feature achieves a PSNR of 24.025 dB and the highest SSIM, whereas fixed output modulation degrades performance. Dynamic modulation at both the value and output reaches the best PSNR of 24.126 dB and LPIPS of 0.113, improving the corresponding Static configuration by 0.721 dB.
\begin{table}[!tb]
    \centering
    \small
    \setlength{\tabcolsep}{4pt}
    \renewcommand{\arraystretch}{0.9}
    \resizebox{\columnwidth}{!}{
        \begin{tabular}{clccc}
            \toprule
            \textbf{Guidance}
            & \textbf{Modulation Position}
            & PSNR$\uparrow$
            & SSIM$\uparrow$
            & LPIPS$\downarrow$ \\
            \midrule
            \multirow{4}{*}{Static}
            & Source Feature        & 24.025 & 0.870 & 0.114 \\
            & Value                 & 24.010 & 0.865 & 0.116 \\
            & Output                & 23.355 & 0.865 & 0.120 \\
            & Both (Value + Output) & 23.405 & 0.863 & 0.119 \\
            \midrule
            \multirow{4}{*}{Dynamic}
            & Source Feature        & 23.575 & 0.866 & 0.117 \\
            & Value                 & 23.551 & 0.863 & 0.120 \\
            & Output                & 23.570 & 0.867 & 0.119 \\
            & Both (Value + Output) & 24.126 & 0.868 & 0.113 \\
            \bottomrule
        \end{tabular}
    }
    \caption{Effects of guidance strategy and modulation position in D2IR on LOLv2-Real.}
    \label{tab:d2ir_ablation}
\end{table}

\section{Conclusion}\label{sec:Conclusion}

This work formulates chrominance contribution conditioned on intensity as a conditional mutual information (CMI) problem and proposes CMIG-Net. CMIC quantifies and calibrates the spatial chrominance contribution, while D2IR combines this prior with restoration states to regulate bidirectional information injection dynamically. Experiments demonstrate improved exposure, color, and detail recovery with less noise and fewer chromatic artifacts. The proposed framework provides an interpretable approach to multi-component interaction and suggests broader applications of CMI in image restoration.


\bibliography{aaai2027}


\end{document}